\documentclass[twoside]{article}

\usepackage[accepted]{aistats2026}
\usepackage[round]{natbib}
\renewcommand{\bibname}{References}
\renewcommand{\bibsection}{\subsubsection*{\bibname}}

\usepackage{hyperref}
\usepackage[capitalise,nameinlink]{cleveref}

\crefname{equation}{Eq.}{Eqs.}
\Crefname{equation}{Equation}{Equations}
\crefformat{equation}{Eq.~#2#1#3}
\Crefformat{equation}{Equation~#2#1#3}
\crefrangeformat{equation}{Eqs.~#3#1#4--#5#2#6}
\crefmultiformat{equation}{Eqs.~#2#1#3}{ and~#2#1#3}{, #2#1#3}{, and~#2#1#3}
\crefname{figure}{Fig.}{Figs.}
\Crefname{figure}{Figure}{Figures}
\crefname{section}{Sec.}{Secs.}
\Crefname{section}{Section}{Sections}
\crefname{table}{Table}{Tables}
\Crefname{table}{Table}{Tables}

\usepackage{graphicx}
\usepackage{algorithm}
\usepackage{algorithmic}
\usepackage{booktabs}
\usepackage{multirow}
\usepackage{xcolor}

\usepackage{amsmath}
\usepackage{amssymb}
\usepackage{mathtools}
\usepackage{amsthm}
\usepackage{tablefootnote}
\usepackage{kotex}

\newcommand{\bx}{\mathbf{x}}

\newcommand{\by}{\mathbf{y}}

\newcommand{\RE}{\ensuremath{\mathbb{R}}}

\theoremstyle{plain}
\newtheorem{theorem}{Theorem}[section]
\newtheorem{proposition}[theorem]{Proposition}

\theoremstyle{definition}

\theoremstyle{remark}

\begin{document}

\runningtitle{Value Gradient Sampler}

\runningauthor{Hwang, Jeong, Shin, Park, Kweon, Yoon, and Park}

\twocolumn[

\aistatstitle{Value Gradient Sampler: Learning Invariant Value Functions \\for Equivariant Diffusion Sampling}

\aistatsauthor{ Himchan Hwang$^{1*}$\quad Hyeokju Jeong$^{1*}$\quad Dong Kyu Shin$^{1*}$\quad Che-Sang Park$^1$\quad Sehee Kweon$^1$ \\\textbf{Sangwoong Yoon}$^{2\dagger}$\quad \textbf{Frank C. Park}$^{1\dagger}$}

\aistatsaddress{ $^1$Seoul National University 
% $^2$Korea Advanced Institute of Science and Technology (KAIST) \\ 
$^2$Ulsan National Institute of Science and Technology (UNIST) \\
$^*$Equal contribution $^\dagger$Corresponding authors} ]

\begin{abstract}
We propose the Value Gradient Sampler (VGS), a diffusion sampler parameterized by value functions. VGS generates samples from an unnormalized target density (i.e., energy) by evolving randomly initialized particles along the gradient of the value function. In many sampling problems where the target density exhibits invariant symmetries, value functions provide a novel approach to leveraging invariant networks for sampling by inducing an equivariant gradient flow, without requiring more complex equivariant networks. The value networks are trained via temporal difference learning, which supports off-policy training and other established reinforcement learning (RL) techniques. By combining advanced RL methods with efficient invariant networks, VGS achieves both the highest sample quality and the fastest sampling speed among our baselines on the 55-particle Lennard-Jones system.
\end{abstract}

\section{Introduction}

Sampling from complex, high-dimensional distributions is a fundamental problem in statistics, machine learning, and the natural sciences. The goal of a sampling algorithm is to generate independent and identically distributed (i.i.d.) samples from an unnormalized target density $q(\bx)$. Recently, samplers inspired by diffusion processes have emerged as a powerful class of algorithms for this task. These methods construct samples via a multi-step diffusion process that interpolates between a simple distribution and the target density, demonstrating strong performance across diverse applications \citep{zhang2022path,vargas2024transport,berner2022optimal,vargas2023denoising,bengio2023gflownetfoundations,lahlou2023theory,havens2025adjoint}.

\begin{figure*}[t]
  \centering
  \begin{minipage}{0.51\textwidth}
    \centering
    \includegraphics[width=0.88\linewidth]{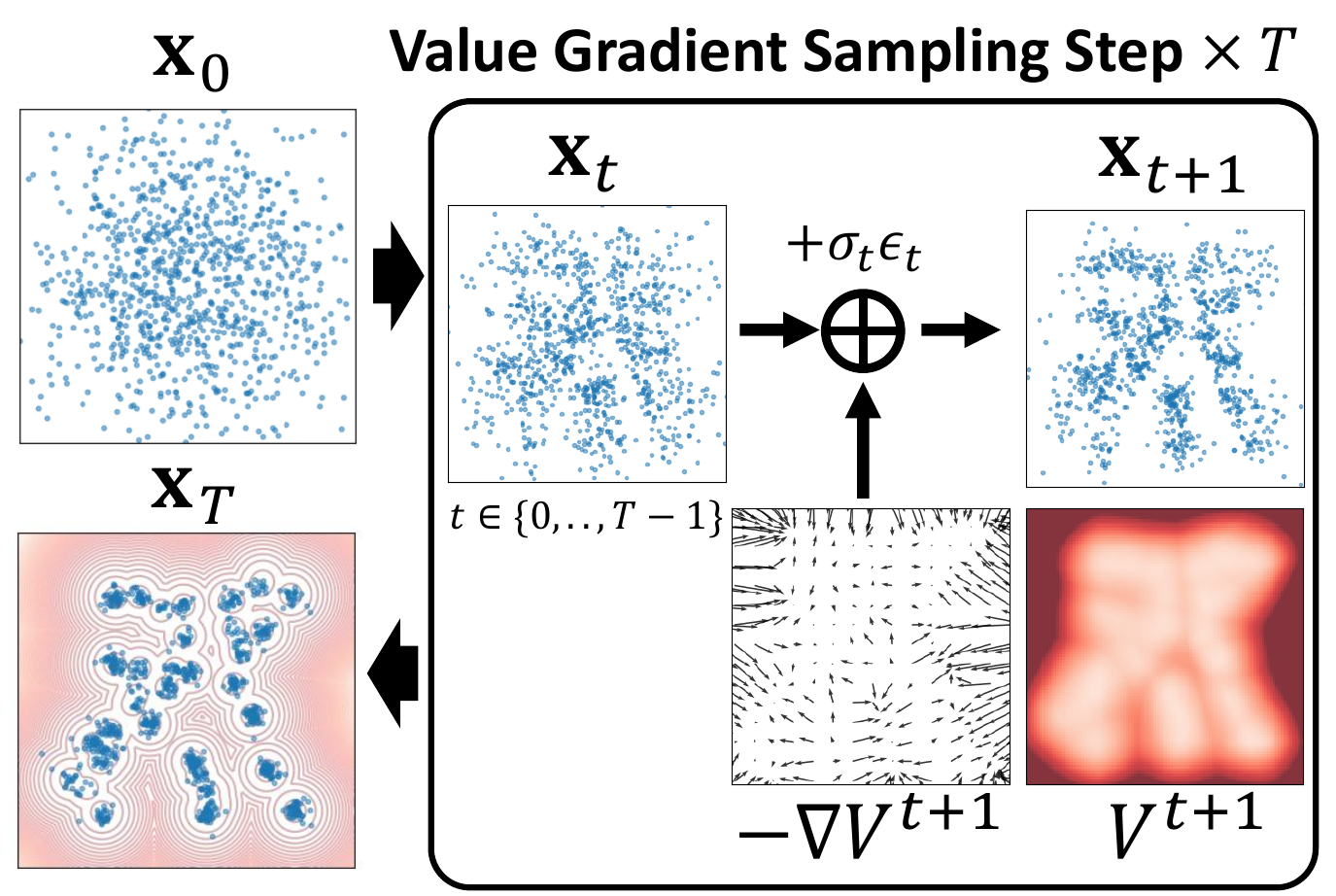}
  \end{minipage}
  \hfill
  \begin{minipage}{0.48\textwidth}
    \centering
    \includegraphics[width=\linewidth]{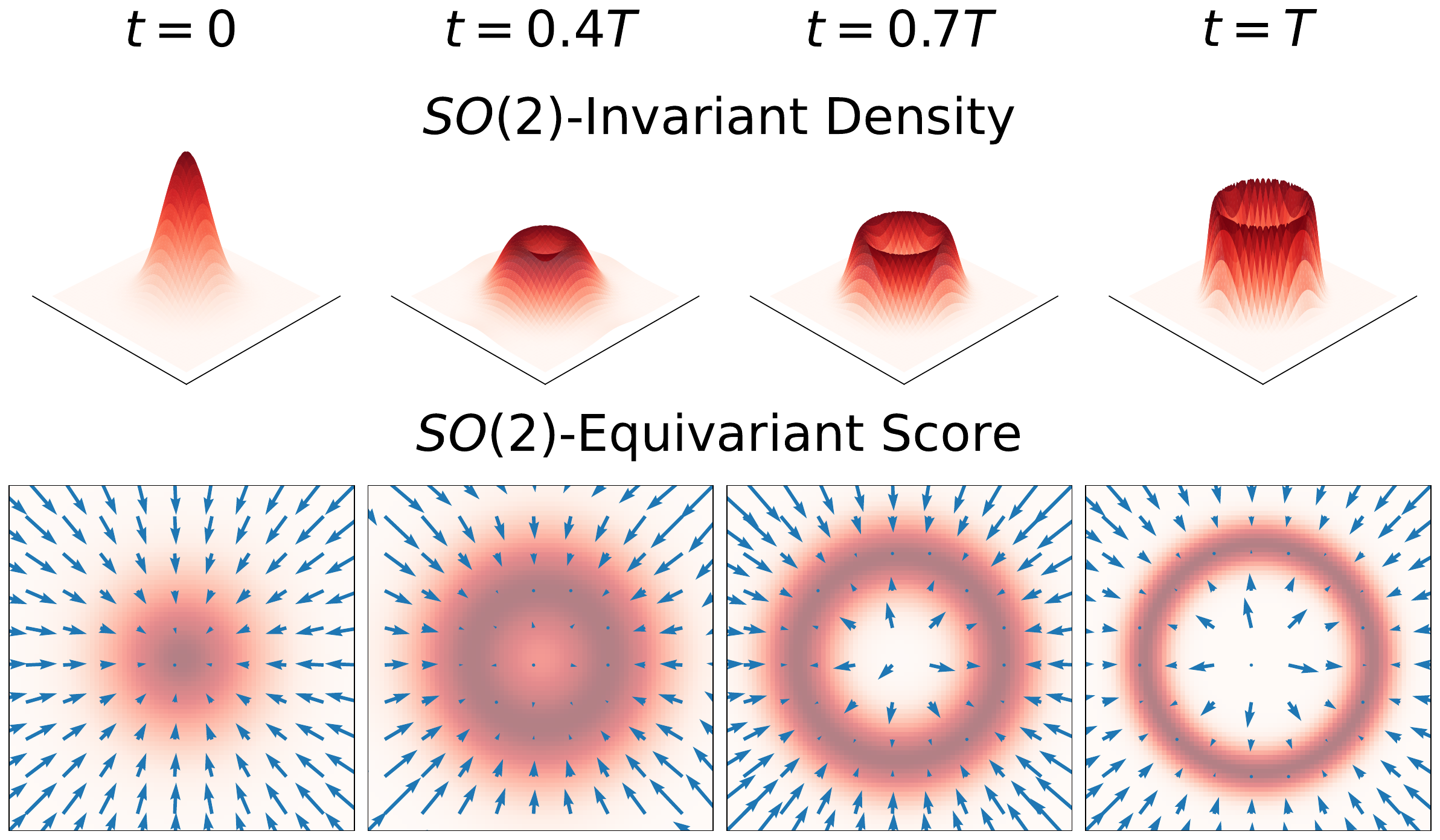}
  \end{minipage}
  \caption{(\textbf{Left}) Value Gradient Sampling. Initial samples are iteratively drifted along the gradient of the next-step value function to match the target distribution. (\textbf{Right}) Diffusion of an $\mathrm{SO}(2)$-invariant target density ($t=T$) via a variance-preserving diffusion process. The diffused densities also exhibit $\mathrm{SO}(2)$-invariance while their score functions possess $\mathrm{SO}(2)$-equivariance. 
  VGS employs a sequence of invariant value functions (which are related to the diffused log-densities) to parameterize the equivariant gradient flow.
  } \label{fig:illustration}
\end{figure*}

In many important applications of sampling, the target density exhibits symmetries.
For instance, the equilibrium state distributions of particle and molecular systems are governed by Euclidean symmetries, such as translation and rotation, which are formalized by the group $\mathrm{SE}(3)$ (or $\mathrm{E}(3)$ if reflection is also considered).
Respecting these symmetries is crucial, as it can lead to improved performance, better sample efficiency, and enhanced generalization \citep{karczewski24generalization}. To sample from an $\mathrm{E}(3)$-invariant target density, most diffusion samplers employ $\mathrm{E}(3)$-equivariant neural network architectures \citep{satorras2021egnn} to parametrize $\mathrm{E}(3)$-equivariant score functions \citep{akhound-sadegh2024iterated,havens2025adjoint}. \looseness=-1

However, enforcing Euclidean equivariance in neural architectures is a non-trivial endeavor. Equivariant networks often rely on mathematically sophisticated constructions, incur significant computational overhead, or impose structural restrictions \citep{kohler2019equivariant,thomas2018tensor,fuchs2020se3transformer,geiger2022e3nn,midgley2023se}. While EGNNs have become widely adopted due to their relatively simple design, they remain slower than general-purpose architectures such as multi-layer perceptrons or graph neural networks \citep{satorras2021egnn}. In particular, the computation in EGNNs inevitably scales as $\mathcal{O}(n^2)$ with respect to the number of particles $n$ in the system due to all-pairs message passing.

In contrast, achieving $\mathrm{E}(3)$-invariance may be considerably simpler, especially for particle systems where the energy depends on the relative positions of particles. A permutation-invariant neural network over pairwise distances and node features can represent an invariant potential without requiring explicitly equivariant layers. Moreover, taking the gradient of such an invariant network naturally induces an equivariant gradient flow \citep{papamakarios2021normalizing}.

In this work, we propose \textbf{Value Gradient Sampler} (VGS), a novel diffusion-based sampling framework that leverages $\mathrm{E}(m)$-invariant rather than $\mathrm{E}(m)$-equivariant neural networks for $m$-dimensional, $n$-particle systems.
Instead of directly parameterizing equivariant score functions, VGS uses a sequence of invariant value functions to represent the diffusion process (\cref{fig:illustration}). These value functions are related to the diffused target densities, which preserve the same symmetries as the original target density. Sampling is then performed by following the gradients of these value functions, which are $\mathrm{E}(m)$-equivariant by construction. In this way, VGS yields an equivariant gradient flow without the need for explicitly equivariant architectures.  \looseness=-1

The value functions in VGS are trained through reinforcement learning (RL). We show that temporal-difference (TD) learning \citep{sutton1988learning} can be naturally applied to minimize an upper bound of the KL divergence between the generated samples and the target density. This connection to RL allows us to incorporate established techniques such as TD($\lambda$), off-policy exploration, and double value networks (\cref{sec:exploration}) to further enhance the performance of VGS.

We demonstrate the effectiveness of VGS on particle system benchmarks, where the goal is to sample equilibrium states given the energy function of a system. VGS achieves the highest sample-quality metrics among the baselines in the LJ-55 setting, a challenging system of 55 indistinguishable particles with an energy function defined in a 165-dimensional space. Simultaneously, VGS dramatically reduces sampling time compared to existing equivariant samplers by replacing a computationally intensive equivariant network with an efficient invariant network. VGS also remains competitive on standard benchmarks without explicit symmetries. Our results highlight that VGS is a promising approach for symmetry-aware sampling applications. The implementation of VGS is publicly available \footnote{\href{https://github.com/swyoon/value-gradient-sampler}{\texttt{https://github.com/swyoon/value-gradient-sampler}}}.\looseness=-1

\section{Preliminaries}

\textbf{Sampling Problems.}\quad
We consider drawing independent samples from a strictly positive target distribution $q(\bx)$ on $\RE^D$ of Boltzmann-form
\begin{align}\label{eq:boltzmann}
    q(\bx)=\frac{1}{Z}\exp{(-E(\bx))},
\end{align}
where $E:\RE^D\to\RE$ is an energy function and $Z=\int_{\RE^D}\exp{(-E(\bx))}d\bx$ is the normalizing constant.
% We adopt a general setting in which $E(\bx)$ can be evaluated pointwise, but the normalizing constant $Z$ and the gradient $\nabla E(\bx)$ are unavailable or prohibitively expensive to compute.
% This arises, for example, when $E(\bx)$ is produced by a black-box simulator or a human preference score, and in high dimensions where $Z$ is intractable.

\textbf{Invariance and Equivariance.}\quad  Given a group $\mathrm{G}$ acting on a set $\mathcal{X}$ via the action $\cdot$, a scalar function $f:\mathcal{X}\to\mathbb{R}$ is \emph{$\mathrm{G}$-invariant} if $f(g\cdot \bx)=f(\bx)$ for all $g\in \mathrm{G}$ and $\bx\in\mathcal{X}$.
A map $F:\mathcal{X}\to\mathcal{X}$ is \emph{$\mathrm{G}$-equivariant} if $F(g\cdot \bx)=g\cdot F(\bx)$ for all $g\in \mathrm{G}$ and $X\in\mathcal{X}$.

\textbf{Symmetry Groups in $n$-Particle Systems.}\quad
Consider the system of $n$ particles in $\mathbb{R}^m$. 
Let the configuration space be $\mathcal{X}=\mathbb{R}^{m\times n}$ with $\bx=[x_1,\ldots,x_n]$, where $x_i\in\mathbb{R}^m$ denotes the coordinates of the $i$-th particle. 
Two symmetry groups act naturally on $\mathcal{X}$: the Euclidean group $\mathrm{E}(m)$ (translations, rotations, reflections) and the symmetric group $\mathrm{S}_n$ (permutations of particle indices).
Elements $(R,p)\in \mathrm{E}(m)$, where $R\in \mathrm{O}(m)$ and $p\in\mathbb{R}^m$, act by
\begin{align}
    (R,p)\cdot \bx = [Rx_1{+}p,\ldots,Rx_n{+}p],
\end{align}
and permutations $\sigma\in \mathrm{S}_n$ act by re-indexing $\sigma \cdot \bx = [x_{\sigma(1)}, ..., x_{\sigma(n)}]$. 
These actions commute, inducing a product action of $\mathrm{E}(m)\times \mathrm{S}_n$ on $\mathcal{X}$. Under this product action, the energy $E(\bx)$ is unchanged, assuming that the particles are indistinguishable. Thus, the energy $E(\bx)$ is $\mathrm{E}(m)\times \mathrm{S}_n$-invariant.

\textbf{Training Parametric Samplers.}\quad
We aim to train a parametric sampler $\pi_\phi(\bx)$ with parameter $\phi$ to generate approximate samples from the target density $q(\bx)$. 
% A parametric sampler can be faster at generating new samples, as computation can be amortized in the training phase.
A natural choice of the objective function is the (reverse) KL divergence between $\pi_\phi(\bx)$ and $q(\bx)$: $\min_\phi \text{KL}(\pi_\phi(\bx)||q(\bx))=\min_\phi \mathbb{E}_{\pi_\phi}\left[\log (\pi_\phi(\bx)/q(\bx))\right],$
which does not require samples from $q(\bx)$ for training. Other objectives, such as log-variance loss, have also been considered \citep{richter2024improved}.

\textbf{Diffusion Samplers.}\quad
Diffusion samplers generate samples through the following iterative drift-diffusion process:
\begin{align}
    \pi(\bx_0) &= \mathcal{N}(0,\sigma_{\text{init}}^2 I), \nonumber \\
    \pi(\bx_{t+1}|\bx_t) &= \mathcal{N}( \alpha_t\bx_t + \mu^t(\bx_t) ,\sigma_t^2I),\label{eq:pls}
\end{align} 
for $t=0,\ldots,T\!-\!1$. The schedules $\{\alpha_t,\sigma_t\}$ and $\sigma_{\text{init}}$ are fixed. In practice, we parameterize the drift as $\mu_\phi^t(\bx_t)$ and thus define the transition $\pi_\phi(\bx_{t+1}|\bx_t)$. Starting from the initial Gaussian sample $\bx_0$, we sequentially scale, drift, and diffuse the sample to obtain the final sample $\bx_T$. We will often write $\bx$ to denote $\bx_T$. The sampler distribution $\pi_\phi(\bx)$ is induced as the marginal distribution of final samples.
This drift-diffusion process (\cref{eq:pls}) is widely found in various methods, such as diffusion models \citep{ho2020ddpm,song2021scorebased}, Euler-Maruyama discretization of SDE samplers \citep{zhang2022path,berner2022optimal,vargas2023denoising}, and continuous GFlowNets \citep{lahlou2023theory}.

\section{Value Gradient Sampling} \label{sec:vgs}

In this section, we introduce \textbf{Value Gradient Sampler} (VGS), a diffusion sampler that parametrizes the drift $\mu_{\phi}^t(\bx_t)$ using a value function $V_{\phi}^t(\bx_t)$. The optimal control formulation and temporal difference training algorithms are described. Incorporating symmetries into VGS will be discussed in \cref{sec:invariant-vgs}.

\subsection{Sampling as Optimal Control}
We train a diffusion sampler $\pi_{\phi}$ to match the target distribution $q$ by leveraging the optimal control view on diffusion-based sampling proposed in \cite{zhang2022path}. Instead of directly minimizing $\text{KL}(\pi_{\phi}(\bx)\Vert q(\bx))$, we minimize an upper bound from the data-processing inequality: $\text{KL}(\pi_{\phi}(\bx)\Vert q(\bx)) \leq \text{KL}(\pi_{\phi}(\bx_{0:T})\Vert \tilde{q}(\bx_{0:T}))$. We set the joint target distribution as
$\tilde{q}(\bx_{0:T})=\tilde{\pi}(\bx_{0:T})q(\bx_T)/\tilde{\pi}(\bx_T)$, which ensures $\tilde{q}(\bx_T) = q(\bx_T)$. Here, $\tilde{\pi}(\bx_{0:T})$ is the joint reference distribution, defined as the uncontrolled process in \cref{eq:pls} with $\tilde{\pi}(\bx_0) = \mathcal{N}(0,\sigma_{\text{init}}^2 I)$ and $\tilde{\pi}(\bx_{t+1}|\bx_t) = \mathcal{N}( \alpha_t\bx_t ,\sigma_t^2I)$.
This construction subsumes prior choices used in PIS \citep[$\sigma_{\text{init}}=0, \alpha_t=1$]{zhang2022path} and DDS \citep[$\sigma_{\text{init}}=1, \alpha_t=\sqrt{1-\sigma_t^2}$]{vargas2023denoising}.

We use a value-based dynamic programming approach to this minimization problem. Minimizing $\text{KL}(\pi(\bx_{0:T})\Vert \tilde{q}(\bx_{0:T}))$ over the admissible class $\Pi$ of policies given by \cref{eq:pls} yields the following optimal control problem:
\begin{align}
    \min_{\pi\in \Pi} \mathbb{E}_{\pi(\bx_{0:T})} \left[
\sum_{t=0}^{T-1} \frac{\Vert\mu^t(\bx_t)\Vert^2}{2\sigma_t^2} + \tilde{E}(\bx_T)
\right],
\end{align}
where $\tilde{E}(\bx_T)=E(\bx_T)+\log \tilde{\pi}(\bx_T)$. See \cref{appendix:objective_dervation} for the derivation. Here, the policy $\pi(\bx_{t+1}|\bx_t)$ is optimized to minimize the sum of running costs $\Vert\mu^t(\bx_t)\Vert^2/{2\sigma_t^2}$, and the terminal cost $\tilde{E}(\bx_T)$.

A value function $V^{t}_{\pi}(\bx_t)$ is the expected sum of future costs starting from $\bx_t$ and following $\pi$:
\begin{align}
V^{t}_{\pi}(\bx_t)=\mathbb{E}_{\pi(\cdot|\bx_t)}\left[
\sum_{i=0}^{T-t-1} \frac{\Vert\mu^{t+i}(\bx_{t+i})\Vert^2}{2\sigma_{t+i}^2} + \tilde{E}(\bx_T) \bigg| \bx_t \right].
\label{eq:value_function}
\end{align}
The optimal value function is the minimum over all possible $\pi(\cdot|\bx_t)$,
$V_*^{t}(\bx_t)=\min_{\pi(\cdot\mid\bx_t)\in \Pi} V^t_{\pi}(\bx_t)$.
If $\tilde{q}(\bx_{t+1:T}|\bx_t)\in\Pi$, then $V_*^{t}(\bx_t)$ is the energy of the marginal density ratio $\tilde{q}(\bx_t)/\tilde{\pi}(\bx_t)$:
\begin{align}\label{eq:opt_val}
    \tilde{q}(\bx_t)/\tilde{\pi}(\bx_t) = \frac{1}{Z}\exp{(-V^t_*(\bx_t))}.
\end{align}
The derivation is in \cref{appendix:opt_val_proof}.
Intuitively, $V_*^{t}(\bx_t)$ is lower on states that are more probable under marginal target density $\tilde{q}(\bx_t)$ compared to the reference density $\tilde{\pi}(\bx_t)$.

% \begin{proposition}[Optimal Value Function] \label{theorem:value}
%     If the admissible set of policies $\pi(\bx_{t+1:T}|\bx_t)$ includes $\tilde{q}(\bx_{t+1:T}|\bx_t)$, the optimal value function $V^t_*(\bx_t)$ is the energy of a marginal density ratio of the target and reference distribution: $\tilde{q}(\bx_t)/\tilde{\pi}(\bx_t) \propto \exp{(-V^t_*(\bx_t))}.$
% \end{proposition}

\begin{algorithm}[t]
      \caption{Value Gradient Sampler} \label{alg:vgs}
      \begin{algorithmic}[1]
        \STATE {\bf Input:} Value $V_{\phi}^t(\bx_t)$, \\ constants $\{\alpha_t\}_{t=0}^{T-1}$, $\{\sigma_t\}_{t=0}^{T-1}$, $\sigma_{\text{init}}$. 
       \STATE $\bx_0 \sim \mathcal{N}(0, \sigma_{\text{init}}^2I)$ \hfill // Initial samples
       \FOR{$t=0$ {\bf to} $T-1$}
       \STATE $\mu_{\phi}^t(\bx_t)=- \sigma_t^2 \textcolor{red}{\nabla_{\alpha_{t}\bx_t} V^{t+1}_{\phi}(\alpha_{t}\bx_t)}$
       \hfill// Eq. (\ref{eq:vgs_sampling})
       \STATE $\epsilon_t \sim \mathcal{N}(0,I)$
       \STATE $\bx_{t+1} = \alpha_t\bx_{t} + \mu_\phi^t(\bx_t) +\sigma_t\epsilon_t$ %\hfill //  Eq. (\ref{eq:pls})
       \ENDFOR
       \STATE {\bf Output:} $\bx_T$ %$\{\bx_t\}_{t=0}^T$
      \end{algorithmic}
\end{algorithm}

\subsection{Value Gradient Sampler}\label{sec:value_grad}
We present \textbf{Value Gradient Sampler} (VGS), a sampler that approximately solves this optimal control problem by drifting a particle along the gradient of the next-step value function.
For our policy at time $t$ defined in \cref{eq:pls},
we will find the drift vector $\mu^t(\bx_t)$ that minimizes the expected future cost from $\bx_t$, which can be represented using the next-step value function $V_{\pi}^{t+1}(\bx_{t+1})$:
\begin{align}
    \min_{\mu^t(\bx_t)} \mathbb{E}_{\pi(\bx_{t+1}|\bx_t)} \left[ \frac{\Vert\mu^t(\bx_t)\Vert^2}{2\sigma_t^2} +V^{t+1}_{\pi}(\bx_{t+1})\bigg| \bx_{t}\right]. \label{eq:vgs-obj}
\end{align}
% For the backward transition, we choose $\tilde{q}(\bx_t|\bx_{t+1})= \mathcal{N}\left(\frac{1}{\alpha_{t}}\bx_{t+1}, s_t^2 I\right)$ with $s_t > 0$ for its mathematical simplicity and connection to diffusion models. Depending on the choice of $\alpha_t$, the process induced by $\tilde{q}(\bx_t|\bx_{t+1})$ becomes equivalent to variance-exploding ($\alpha_t = 1$) or variance-preserving ($\alpha_t = 1/\sqrt{1-s_t^2}$) diffusion processes.
% Now we can write the optimization for $\mu_t$ and $\sigma_t$ as follows: \looseness=-1
% \begin{align}
%     \min_{\mu_t, \sigma_t} 
%     \mathbb{E}_{\epsilon}
%          \left[V^{t+1}_{\pi}(\alpha_{t}\bx_{t}+\mu_t+\sigma_t\epsilon)\right] - \tau D \log \frac{\sigma_t}{s_{t}}
%     +
%     \frac{\tau ||\mu_t||^2 }{2s_t^2 \alpha_{t}^{2}} + \frac{\tau D\sigma_t^2}{2s_t^2 \alpha_{t}^{2}}-\tau\frac{D}{2},  \label{eq:optimal_policy}
% \end{align}
As this objective lacks a closed-form solution, we apply a first-order Taylor expansion to $V^{t+1}_{\pi}(\bx_{t+1})$ around $\alpha_t\bx_t$ to obtain:
% Unfortunately, an analytical solution to this problem is not available. In search of an approximation for optimal $\mu_t$ and $\sigma_t$, we use a first-order Taylor expansion of $V^{t+1}_{\pi}$ with respect to $\alpha_{t}\bx_{t}$. 
\begin{align}
    \mathop{\mathbb{E}}_{\pi(\bx_{t+1}|\bx_t)}& \left[ V^{t+1}_{\pi}(\bx_{t+1})\right] =
    \mathbb{E}_{\epsilon_t}[V^{t+1}_{\pi}(\alpha_{t}\bx_{t}+\mu^t+\sigma_t\epsilon_t)] \nonumber \\
    &\approx V_{\pi}^{t+1}(\alpha_{t}\bx_{t}) + \mu^t(\bx_t)^\top \nabla_{\alpha_{t}\bx_{t}} V^{t+1}_{\pi}(\alpha_{t}\bx_t), \nonumber
    % \label{eq:vgs-obj-taylor} 
\end{align}
where $\epsilon_t \sim \mathcal{N}(0,I)$. Substituting this approximation back into \cref{eq:vgs-obj} and setting the derivative with respect to $\mu^t$ to zero yields the optimal drift:
\begin{align}
    \mu^t(\bx_t) = - \sigma_t^2 \nabla_{\alpha_{t}\bx_{t}} V^{t+1}_{\pi}(\alpha_{t}\bx_t). \label{eq:vgs_sampling}
\end{align}
The detailed derivation is presented in \cref{appendix:second_order}.
Note that $\mu^t(\bx_t)$ is now determined entirely by the gradient of the next-step value function. In practice, we use a parameterized value network $V^t_\phi(\bx_t)$ mapping $(\bx_t, t)$ to a scalar, substituting $V_\phi$ for $V_\pi$ in \cref{eq:vgs_sampling} for sampling (\cref{alg:vgs}). To approach the optimal value function and policy, we follow the framework of generalized policy iteration, which alternates between policy evaluation and improvement. VGS corresponds to the policy improvement step, refining the policy based on the current value function. The policy evaluation step is described in the next section.

\subsection{Temporal Difference Learning}\label{sec:learn_value}
We employ TD learning \citep{sutton1988learning} to estimate the value function from samples. From the definition of the value function (\cref{eq:value_function}), we obtain the following recurrence relation:
$V^{t}_{\pi}(\bx_t)=\mathop{\mathbb{E}}_{\pi(\bx_{t+1}|\bx_{t})}\left[\Vert\mu^t(\bx_t)\Vert^2/2\sigma_t^2+V^{t+1}_{\pi}(\bx_{t+1}) | \bx_t \right]$. To approximate $V^t_\pi$, we train a parameterized value network $V^t_\phi$ to satisfy this recurrence, resulting in a regression-type loss with the TD target $\hat{V}^t_{\text{TD}}(\bx_{t:t+1})$.
The TD target is defined using a target network $V_{\phi^-}^t(\bx_t)$, a detached copy of the value network parametrized by $\phi^-$. 
Specifically, given a transition $\bx_{t+1}\sim \pi_{\phi^-}(\bx_{t+1}|\bx_t)$, we minimize the mean squared TD error:
\begin{align}
\raisetag{6pt} 
\min_{\phi}\; \mathbb{E}_{\pi_{\phi^-}(\bx_{t+1}\mid \bx_t)}
&\left[(V_{\phi}^{t}(\bx_t)-\hat V^t_{\text{TD}}(\bx_{t:t+1}))^2\right],
\label{eq:td_error_minimization}\\
\text{where}\quad
\hat V^t_{\text{TD}}(\bx_{t:t+1})
&= \frac{\|\mu_{\phi^-}^{\,t}(\bx_t)\|^2}{2\sigma_t^2}
+ V_{\phi^-}^{\,t+1}(\bx_{t+1}).\notag
\end{align}
Gradients are not backpropagated into the target network parameters $\phi^-$. Instead, $\phi^-$ is updated separately via exponential moving averaging: $\phi^{-} \leftarrow \kappa \phi^- + (1-\kappa)\phi$ for $0<\kappa<1$. This target-network scheme is known to stabilize training with function approximation \citep{mnih2015human,haarnoja18soft}. The overall training procedure is summarized in \cref{alg:vgs_training} in the Appendix.

Note that there exist critical differences between our TD learning and Detailed Balance (DB) \citep{bengio2023gflownetfoundations}. First, we employ a fixed target network during the update (\cref{eq:td_error_minimization}). Second, the TD loss does not directly update the policy. Instead, the policy is implicitly optimized via taking the gradient of the next-step value (\cref{eq:vgs_sampling}). Consequently, only the current state value is updated based on the future state value, allowing information from terminal states to propagate to earlier states in a stable and rectified manner.

\textbf{TD($\lambda$).}\quad 
The TD learning described above, often referred to as TD(0), can be slow in propagating information in time, as only a single-step transition is used in the update.
Alternatively, TD($\lambda$) \citep{sutton1988learning} incorporates all available future temporal differences by computing a weighted average with exponentially decaying weights.
We first define the TD error as follows:
\begin{align}
\raisetag{6pt} 
    \delta_t(\bx_t, \bx_{t+1}) =  \frac{\Vert\mu^t_{\phi^-}(\bx_t)\Vert^2}{2\sigma_t^2} + V^{t+1}_{\phi^-}(\bx_{t+1}) - V^{t}_{\phi^-}(\bx_{t}). \label{eq:delta_t} 
    % \nonumber
\end{align}
% The TD target in \cref{eq:td_target_net} is a biased estimate of $V^t_\phi$, as the unknown $V_{\pi}^{t+1}(\bx_{t+1})$ is replaced with its parametric approximation $V_{\phi}^{t+1}(\bx_{t+1})$. 
% In contrast, an unbiased Monte Carlo estimate suffers from high variance. 
% TD($\lambda$) offers a principled trade-off between bias and variance by interpolating between these two extremes. 
% When $\lambda = 0$, TD($\lambda$) target is reduced to the TD target in \cref{eq:td_target_net}, and when $\lambda = 1$, it recovers the Monte Carlo estimate using the full trajectory. 
Given a trajectory $\bx_{t+1:T} \sim \pi_{\phi^-}(\bx_{t+1:T}|\bx_t)$ and a constant $\lambda \in [0, 1]$, the TD($\lambda$) target is defined as:
\begin{align}
\raisetag{6pt} 
    \hat{V}^t_{\text{TD}(\lambda)}(\bx_{t:T}) = V_{\phi^-}^{t}(\bx_t) + \sum_{i=0}^{T-t-1}\lambda^{i}\delta_{t+i}(\bx_{t+i}, \bx_{t+i+1}), \label{eq:td_lambda_target}
\end{align}
which can be used in \cref{eq:td_error_minimization} instead of $\hat{V}^t_{\text{TD}}(\bx_{t:t+1})$. This formulation offers a principled trade-off between bias and variance, interpolating between one-step TD ($\lambda=0$) and Monte Carlo estimation ($\lambda=1$).

SubTB($\lambda$) \citep{madan2023learning} implements a similar idea but applies a weighted average over subtrajectory losses, whereas TD($\lambda$) incorporates it directly into the target calculation. By looking ahead multiple steps, TD($\lambda$) is often more efficient and effective than TD(0) in assigning credit to earlier actions. This benefit becomes more pronounced for larger $T$, as demonstrated in our comparison experiments in \cref{sec:synthetic_exp}.

\subsection{Leveraging RL Techniques in VGS} \label{sec:exploration}
The TD formulation enables us to leverage well-established techniques in deep RL.
% Since training VGS is essentially a value-based RL, we can leverage well-established techniques in deep RL. Here, we introduce the methods used in our experiments. We provide ablations in  \cref{sec:particle_exp}.

\textbf{Training with an Exploration Policy.}\quad 
An exploration policy $\pi_{\text{expl}}$ is often used in RL to promote exploration. 
Assuming the transition density $\pi_{\text{expl}}(\bx_{t+1}|\bx_t)$ is tractable, off-policy TD($\lambda$) targets can be computed on a sampled trajectory $\bx_{0:T} \sim \pi_{\text{expl}}(\bx_{0:T})$ using importance sampling. 
The TD targets for all timesteps $t = 0,\dots,T$ can be computed from the stored tuples $\{\bx_t, \mu^t_{\phi^-}(\bx_t)\}_{t=0}^T$, collected during the sampling phase.
Details on off-policy TD($\lambda$) computation are given in \cref{appendix:off_td_lambda}.
In our experiments, we set the exploration policy as the current policy with an amplified noise level $(\sigma_t)_{\text{expl}} = \eta \sigma_t$ for a scale factor $\eta > 1$, following prior works \citep{lahlou2023theory, madan2023learning, sendera2024improved}.

\textbf{Double Value Networks.}\quad Using a single value function for both target evaluation and action optimization induces an overestimation bias in RL \citep{thrun2014issues}. TD3 addresses this issue by maintaining two independent value networks: the policy is optimized using one critic, while targets are computed as the minimum of the two critics’ estimates \citep{fujimoto2018addressing}. In VGS, we adopt the same principle: using a single-value network for the value-gradient sampling (policy optimization) step and computing TD targets as the minimum of two value estimates. This design increases training time computation but leaves the sampling cost unchanged. We empirically found that the double-value approach improves training stability and convergence for VGS.

% In practice, we store tuples $(\bx_t, \mu^t_\phi(\bx_t))$ during the sampling phase, and use it to compute the it  recomputing the running cost $\tau\log{\pi_\phi(\bx_{t+1}|\bx_t)/\tilde{q}(\bx_t|\bx_{t+1})}$ and the importance weight $\pi_\phi(\bx_{t+1}|\bx_t) / \pi_{\text{expl}}(\bx_{t+1}|\bx_t)$.

\textbf{Training with Sample Buffers.}\quad In some cases, important samples $\bx_T$ may be externally provided, such as MCMC samples, or states selected from a prioritized replay buffer. To leverage such data, we sample the backward trajectory $\bx_{0:T-1} \sim \tilde{q}(\bx_{0:T-1}|\bx_T)$ and perform TD updates along the resulting trajectory. We prefer TD(0) in this setting, as TD($\lambda$) requires computationally expensive resampling of full future trajectories $\bx_{t+1:T} \sim \pi_{\phi^-}(\bx_{t+1:T}|\bx_t)$ at each step, whereas TD(0) efficiently uses a single transition. We evaluate VGS with sample buffers in our synthetic distribution experiments in \cref{sec:synthetic_exp}.
Following \cite{sendera2024improved}, we alternate training VGS on forward and backward trajectories.

\begin{table*}[t]
    \centering
    \caption{Results of $n$-body particle system experiments. The metrics are evaluated with 10,000 samples. We report the average and standard deviation over three random seeds. Metrics shown in \textbf{bold} denote the best mean performance, while \underline{underlined} entries indicate means are not statistically distinguishable from the best mean, using a one-sided Welch's t-test with a significance threshold of $p < 0.1$. $*$ indicates divergent training. For details, see \cref{appendix:particle_detail}.} 
    \setlength{\tabcolsep}{0.5pt}
    % \scriptsize
    \resizebox{1\linewidth}{!}
    {\begin{tabular}{lccccccccc}
        \toprule
         Energy & \multicolumn{3}{c}{DW-4 ($D$=8)} & \multicolumn{3}{c}{LJ-13 ($D$=39)} & \multicolumn{3}{c}{LJ-55 ($D$=165)} \\
         \cmidrule(lr){1-1}\cmidrule(lr){2-4}\cmidrule(lr){5-7}\cmidrule(lr){8-10}
          Metric & TVD-D & TVD-E & $\mathcal{W}_{2}$ & TVD-D & TVD-E & $\mathcal{W}_{2}$ & TVD-D & TVD-E & $\mathcal{W}_{2}$ \\
         \midrule
         FAB 
         & 0.073\tiny{$\pm$0.004}
         & 0.154\tiny{$\pm$0.011}
         & 1.931\tiny{$\pm$0.063}
         & \underline{0.278\tiny{$\pm$0.072}}
         & 0.950\tiny{$\pm$0.044}
         & 5.350\tiny{$\pm$0.231}
         & 0.148\tiny{$\pm$0.006}
         & \underline{1.000\tiny{$\pm$0.000}}
         & 17.217\tiny{$\pm$0.466} \\
         iDEM
         & \underline{0.061\tiny{$\pm$0.003}}
         & 0.118\tiny{$\pm$0.013}
         & 1.597\tiny{$\pm$0.011}
         & \underline{0.029\tiny{$\pm$0.005}}
         & 0.154\tiny{$\pm$0.024}
         & \textbf{3.976\tiny{$\pm$0.004}}
         & \underline{0.160\tiny{$\pm$0.104}}
         & \underline{0.951\tiny{$\pm$0.086}}
         & \underline{16.460\tiny{$\pm$1.123}} \\
        DiKL
        & \underline{0.061\tiny{$\pm$0.009}}
        & 0.193\tiny{$\pm$0.079}
        & \textbf{1.566\tiny{$\pm$0.013}}
        & 0.070\tiny{$\pm$0.016}
        & 0.488\tiny{$\pm$0.111}
        & 4.058\tiny{$\pm$0.035}
        & 0.279\tiny{$\pm$0.121}
        & \underline{1.000\tiny{$\pm$0.000}}
        & \underline{18.050\tiny{$\pm$2.158}} \\
         \midrule
         PIS 
         & 0.144\tiny{$\pm$0.015}
         & 0.388\tiny{$\pm$0.014}
         & 2.107\tiny{$\pm$0.120}
         & 0.285\tiny{$\pm$0.069}
         & 0.523\tiny{$\pm$0.045}
         & \underline{4.187\tiny{$\pm$0.296}}
         & $*$
         & $*$
         & $*$ \\
         DDS
         & 0.259\tiny{$\pm$0.024}
         & 0.573\tiny{$\pm$0.062}
         & 3.327\tiny{$\pm$0.250}
         & $*$
         & $*$
         & $*$
         & $*$
         & $*$
         & $*$\\
         GFN-DB
         & 0.476\tiny{$\pm$0.005}
         & 0.891\tiny{$\pm$0.092}
         & 2.739\tiny{$\pm$0.938}
         & $*$
         & $*$
         & $*$
         & $*$
         & $*$
         & $*$ \\
         GFN-SubTB
         & 0.218\tiny{$\pm$0.007}
         & 0.585\tiny{$\pm$0.011}
         & 1.606\tiny{$\pm$0.006}
         & 0.385\tiny{$\pm$0.007}
         & 0.999\tiny{$\pm$0.001}
         & 5.210\tiny{$\pm$0.036}
         & $*$
         & $*$
         & $*$ \\
         GFN-TB
         & 0.350\tiny{$\pm$0.021}
         & 0.743\tiny{$\pm$0.020}
         & 1.687\tiny{$\pm$0.046}
         & 0.325\tiny{$\pm$0.030}
         & 0.982\tiny{$\pm$0.002}
         & 4.914\tiny{$\pm$0.185}
         & $*$
         & $*$
         & $*$ \\
         \midrule
         VGS (Ours)
         & \textbf{0.053\tiny{$\pm$0.008}}
         & \textbf{0.097\tiny{$\pm$0.016}}
         & 1.587\tiny{$\pm$0.007}
         & \textbf{0.025\tiny{$\pm$0.001}}
         & \textbf{0.109\tiny{$\pm$0.024}}
         & 4.029\tiny{$\pm$0.011}
         & \textbf{0.061\tiny{$\pm$0.023}}
         & \textbf{0.799\tiny{$\pm$0.297}}
         & \textbf{15.906\tiny{$\pm$0.261}} \\
         \bottomrule
    \end{tabular}}
    \label{tab:particle_tab}
    \vskip -0.2cm
\end{table*}

\section{Value Gradient Sampler for \texorpdfstring{$n$}{n}-Particle Systems} \label{sec:invariant-vgs}

A sampler that generates equilibrium states of a many-body system is called a Boltzmann generator \citep{noe2019boltzmann}. In addition to being a challenging sampling benchmark, inferring stable configurations given an energy function is an important scientific problem in chemistry and biology \citep{jumper2021highly, hoogeboom2022equivariant,havens2025adjoint}.
The energy of an $n$-body system is invariant to translation, rotation, reflection, and permutation of particles, exhibiting $\mathrm{E}(m) \times \mathrm{S}_n$-invariance. 
To respect this symmetry, previous approaches have built samplers using $\mathrm{E}(m)$-\emph{equivariant} networks \citep{akhound-sadegh2024iterated, he2024training,midgley2023se}.
In this section, we describe how VGS is applied to the task of sampling equilibrium states of $n$-particle systems. We formulate the problem on a zero-mean space, establish the invariance of the value function, and thus parameterize it using $\mathrm{E}(m)$-\emph{invariant} networks.

\subsection{Zero-mean Space}
The configuration of a system $\bx$ can be projected onto the zero-mean subspace $\mathcal{X}$, utilizing the translation invariance of the system. 
The zero-mean space $\mathcal{X} = \{\bx \in \mathbb{R}^{m\times n} \mid \sum_{i=1}^{n} x_i = 0\}$ is an $m(n\!-\!1)$-dimensional subspace of $\mathbb{R}^{m\times n}$ where the particles satisfy the zero-mean constraint. 
% A and consider the data to be within this subspace.
As $\mathcal{X}$ is a vector space isomorphic to $\mathbb{R}^{m(n\!-\!1)}$, the arguments in \cref{sec:vgs} remain valid, except that the effective dimension reduces to $D_{\text{eff}}=m(n\!-\!1)$. Working in $\mathcal{X}$ also makes the Boltzmann distribution (\cref{eq:boltzmann}) well defined, as the normalizing constant becomes finite. In practice, we project both the drift $\mu_{\phi}^t$ and the noise $\epsilon_t$ onto $\mathcal{X}$ at each step. 

\subsection{Invariance of the Value Function}
On the zero-mean space $\mathcal{X}$, translations are removed, so an $\mathrm{E}(m)\times \mathrm{S}_n$–invariant energy $E$ becomes $\mathrm{O}(m)\times \mathrm{S}_n$–invariant. The main claim of this subsection is that the value function of VGS $V_\pi^t$ and the optimal value function $V_*^t$ inherit exactly the same $\mathrm{O}(m)\times \mathrm{S}_n$ invariance. Let $\circ$ denote the product action of $\mathrm{O}(m)\times \mathrm{S}_n$ on $\mathcal{X}$: $(R, \sigma)\circ\bx = [Rx_{\sigma(1)}, ..., Rx_{\sigma(n)}]$. 

\begin{proposition}[Invariance of $V_\pi^t$ and $V_*^t$]
\label{proposition:invariance}
    Assume that the energy function is $\mathrm{O}(m)\times \mathrm{S}_n$–invariant as follows:
    \begin{align}
        E((R, \sigma) \circ \bx) =  E(\bx) \quad \forall \bx\in \mathcal{X}, (R, \sigma)\in \mathrm{O}(m)\times \mathrm{S}_n. \nonumber
    \end{align}
    then both the value function of VGS and the optimal value function preserves $\mathrm{O}(m)\times \mathrm{S}_n$-invariance:
    \begin{align}
        V^{t}_{\pi}((R, \sigma) \circ \bx_t) =  V^t_{\pi}(\bx_t), \quad 
        V^{t}_{*}((R, \sigma) \circ \bx_t) =  V^t_{*}(\bx_t), \nonumber
    \end{align}
    for all $\bx_t\in \mathcal{X}$, $(R, \sigma)\in \mathrm{O}(m)\times \mathrm{S}_n$.
\end{proposition}
\noindent\textit{Proof sketch.}
The claim for $V_\pi^t$ follows by backward induction from $t=T$ to $0$, using that the gradient of a $\mathrm{G}$–invariant function is $\mathrm{G}$–equivariant when $\mathrm{G}$ acts orthogonally on $\mathcal{X}$ \citep[Lemma~2]{papamakarios2021normalizing}. The statement for $V_*^t$ follows from \cref{eq:opt_val}. See \cref{appendix:thm4.1} for details.

An analogous result holds when particles are distinguishable and the symmetry reduces to $\mathrm{E}(m)$–invariance. The corresponding proposition and proof are also given in \cref{appendix:thm4.1}

The results of this section establish $\mathrm{O}(m)\times \mathrm{S}_n$ invariance on $\mathcal{X}$, which lifts to $\mathrm{E}(m)\times \mathrm{S}_n$ invariance on the full space $\mathbb{R}^{m\times n}$.

\subsection{Invariant Architectures for VGS}
Building on these theoretical results, we consider two invariant architectures in our experiments: an $\mathrm{E}(m)$-invariant graph neural network (IGNN) and an $\mathrm{E}(m)\times \mathrm{S}_n$-invariant multilayer perceptron (IMLP). IGNN can be scaled to general systems when only $\mathrm{E}(m)$ invariance is present. IMLP is a simple and efficient choice when both $\mathrm{E}(m)$ and $\mathrm{S}_n$ invariances hold.

\textbf{\boldmath $\mathrm{E}(m)$-Invariant Graph Neural Network.}\quad
IGNN is a simplification of EGNN, introduced in \cite{satorras2021egnn}. For particles $x_i\in\mathbb{R}^m$, let the pairwise distance be $d_{ij}=\lVert x_i-x_j\rVert$. Given initial node embeddings $h^0=\{h_i^0\}_{i=1}^n$, a layer updates as follows:
\begin{align}
m_{ij}&=\Phi_e(h_i^l,h_j^l,d_{ij}),\qquad
h_i^{l+1}=\Phi_h\left(h_i^l,\sum_{j\ne i} m_{ij}\right). \nonumber
\end{align}
Aggregating the final-layer node embeddings and applying MLPs yields an $\mathrm{E}(m)$-invariant output. To handle distinguishable particles, per-particle features (e.g., atom types) can be embedded in $h^0$. In our experiments where particles are indistinguishable, we use a uniform $h^0$ for both EGNN and IGNN, resulting in $\mathrm{E}(m)\times \mathrm{S}_n$ equivariance and invariance, respectively. A limitation of this architecture is the $\mathcal{O}(n^2)$ computational scaling induced by all-pairs message passing.

\textbf{\boldmath $\mathrm{E}(m)\times\mathrm{S}_n$-Invariant Multilayer Perceptron.}\quad
IMLP is an MLP that takes sorted pairwise distances $\{d_{ij}\}_{i<j\le n}$ as input instead of particle coordinates $\bx$. By construction, the input representation, and thus the output of the network, is $\mathrm{E}(m)\times \mathrm{S}_n$ invariant. Furthermore, replacing $\bx$ with the sorted distances incurs no loss of information up to an $\mathrm{E}(m)\times \mathrm{S}_n$ transformation under mild conditions. This follows from \citep[Theorem 2.6]{boutin2004reconstructing}, stating that the distance multiset almost always reconstructs the spatial configuration if $n\ge m+2$, a condition satisfied across all of our evaluated benchmark systems. Although the input dimension scales as $\mathcal{O}(n^2)$, the forward cost is governed mainly by network width and depth. In the LJ-13 to LJ-55 setting, we observed that it is sufficient to only double the hidden dimension, even though the input grew by about $19\times$. This computational advantage is reflected in the sampling time experiment results in \cref{sec:particle_exp}.

\section{Experiments}\label{sec:experiment}
We evaluate VGS on two tasks: sampling equilibrium states of $n$-body systems (\cref{sec:particle_exp}) and sampling from synthetic distributions (\cref{sec:synthetic_exp}). In this section, we provide a brief overview of the experimental setups and present the main results. Detailed descriptions of the target distributions are provided in \cref{appendix:target_dist}, and the evaluation metrics are provided in \cref{appendix:metrics}. Additional experimental details and hyperparameters are given in \cref{appendix:exp_setup}.
 %\cref{appendix:experiment_appendix}
%For additional details, refer to \cref{appendix:synthetic_details}.

{
\setlength{\tabcolsep}{2pt}
\begin{table}[t]
    % \vskip -0.25cm
    \centering
    \caption{Sampling time measured in the LJ-55 experiments. We measured the average time required for batch size 512 sample generation over 10 trials on a single 24GB RTX 4090 GPU.}
    %\cref{appendix:metrics} and \cref{appendix:synthetic_details}
    \label{tab:sampling_time}
    % \small 
    \resizebox{1\linewidth}{!}{
    \begin{tabular}{lcccc}
        \toprule
          &  $T$ & time/step (ms) & time (s) & \# Param. \\
          \midrule
         FAB & 1\tablefootnote{FAB uses 1 step normalizing flow followed by multiple MCMC steps}  & 5907.94 & 5.91\scriptsize{$\pm$0.51} & 5.44M\\
         iDEM & 1000 & 135.39 & 135.39\scriptsize{$\pm$0.04} & 0.58M\\
         DiKL  & 1 & 368.41 & 0.37\scriptsize{$\pm$0.00} & 2.08M\\ 
         PIS & 200 & 136.91 & 27.38\scriptsize{$\pm$0.05} & 0.58M\\
         DDS & 200 & 136.91 & 27.38\scriptsize{$\pm$0.05} & 0.58M\\
         GFN\tablefootnote{Includes GFN-DB, GFN-TB, and GFN-SubTB.} & 100 & 134.51 & 13.45\scriptsize{$\pm$0.01} & 0.58M\\
         VGS-IMLP & 100 & \textbf{1.28}& \textbf{0.13\scriptsize{$\pm$0.00}} & 10.05M\\
         VGS-IGNN & 100 & 124.43 & 12.43\scriptsize{$\pm$0.01} & 0.32M\\
         \bottomrule
    \end{tabular}
    }
    \vskip -0.2cm
\end{table}
}

\subsection{Sampling \texorpdfstring{$n$}{n}-Body Particle Systems}\label{sec:particle_exp}

\textbf{Target Distributions.}\quad 
We evaluate our method on three standard benchmark $n$-body systems: DW-4, LJ-13, and LJ-55 \citep{kohler2020equivariantflowsexactlikelihood}. 
DW-4 is a 2D 4-particle system with a double-well potential, while LJ-13 and LJ-55 are 3D systems with 13 and 55 particles interacting via the Lennard-Jones potential. 
The input space dimensionality of DW-4, LJ-13, and LJ-55 is 8, 39, and 165, respectively. The potential function of DW-4 exhibit $\mathrm{E}(2)\times \mathrm{S}_4$-invariance, while the potentials of LJ-13 and LJ-55 have $\mathrm{E}(3)\times \mathrm{S}_n$-invariance ($n=13,55$).

\textbf{Performance Metrics.}\quad 
Following \cite{he2024training}, we use three metrics: total variation distance of interatomic distances (TVD-D~$\downarrow$), total variation distance of energy (TVD-E~$\downarrow$), and Wasserstein-2 distance ($\mathcal{W}^2~\downarrow$) between test and generated samples. TVD-D and TVD-E are computed from histograms of interatomic distances and energy, respectively. 
Samples are normalized to zero mean when computing $\mathcal{W}^2$.

\textbf{Baselines.}\quad 
Baselines include FAB with an $\mathrm{SE}(3)$-augmented coupling flow \citep{midgley2023flow, midgley2023se}, iDEM \citep{akhound-sadegh2024iterated}, and DiKL \citep{he2024training}, all explicitly designed for particle system benchmarks by leveraging symmetry. Additionally, we include PIS \citep{zhang2022path} and DDS \citep{vargas2023denoising} as implemented in \cite{akhound-sadegh2024iterated}. We also evaluate GFlowNet (GFN) variants, including GFN-DB \citep{bengio2023gflownetfoundations}, GFN-TB \citep{lahlou2023theory}, and GFN-SubTB \citep{zhang2023diffusion}, using the codebase from \cite{sendera2024improved}. However we exclude their proposed variant (GFN-TB-Imp in \cref{sec:synthetic_exp}) as its MCMC exploration caused divergence. For these additional baselines, we use an EGNN-parametrized equivariant policy configured identically to iDEM. We omit adjoint sampling \citep{havens2025adjoint} due to the lack of publicly available code for the particle systems.

\textbf{Results.}\quad Experimental results are shown in \cref{tab:particle_tab}. VGS outperforms all baseline methods across all particle systems in terms of the TVD-D and TVD-E metrics. For the $\mathcal{W}^2$ metric, DiKL and iDEM achieve the best performance on DW-4 and LJ-13, respectively, while VGS records the lowest average $\mathcal{W}^2$ on LJ-55, demonstrating its scalability to higher dimensions. The results for VGS are obtained using IMLP, double value functions, and an exploration policy. We set $T=100$ and $\lambda=0.9$ for LJ-13 and LJ-55, and $T=50$ and $\lambda=0$ for DW-4. Sample buffers are not used for particle system experiments.

Consistent with the findings of \cite{akhound-sadegh2024iterated}, the training of PIS and DDS tends to diverge in higher dimensions. Similar divergent behavior is also observed in the GFN baselines, which, to the best of our knowledge, we are the first to evaluate on particle system benchmarks. We consider a sampler to have diverged if its TVD-D metric is higher than that of an untrained sampler. Despite explicitly incorporating symmetry via the EGNN, these additional baselines yielded poor performance.

\begin{figure}[t]
    \centering
    \includegraphics[width=1.0\linewidth]{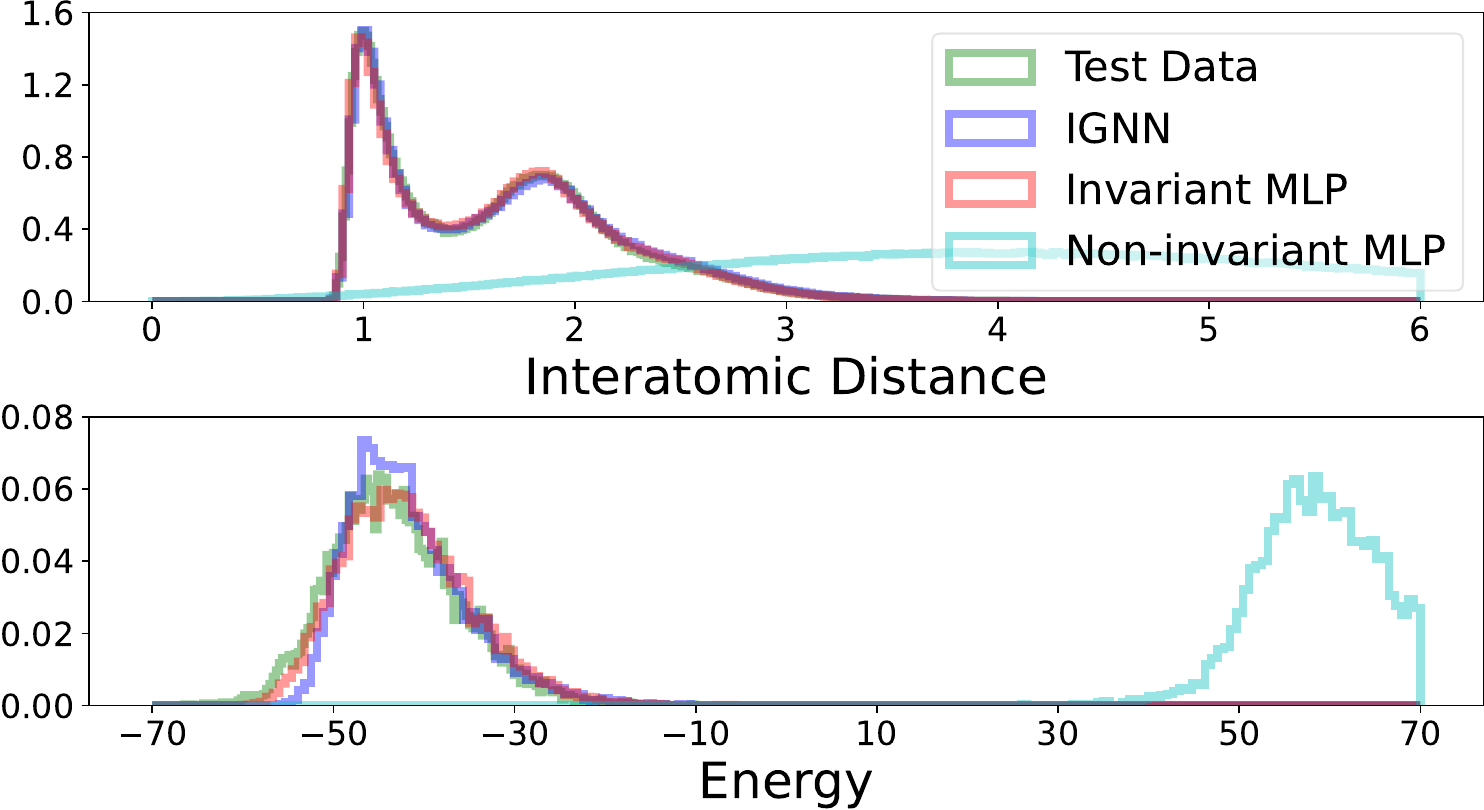}
    \vskip -0.2cm
    \caption{Interatomic Distance (\textbf{Top}) and Energy (\textbf{Bottom}) histograms from LJ-13. Samples from VGS with an invariant MLP, IGNN, and a non-invariant MLP are compared against the test samples. VGS generates accurate samples with invariant networks.}
    \label{fig:lj13_histogram}   
\end{figure}

{
\setlength{\tabcolsep}{2pt}
\begin{table}[t]
    \vskip -0.2cm
    \centering
    \caption{Ablation study on training techniques in VGS. Experiments are conducted in the LJ-13 system.} 
    %\cref{appendix:metrics} and \cref{appendix:synthetic_details}
    \label{tab:ablation}
    % \small
    \resizebox{1\linewidth}{!}{
    \begin{tabular}{lccc}
        \toprule
          &  TVD-D & TVD-E & $\mathcal{W}_2$ \\
          \midrule
         VGS & 0.025\tiny{$\pm$0.00} & 0.109\tiny{$\pm$0.02} & 4.029\tiny{$\pm$0.01}\\ 
         $-$ Double Value & 0.023 & 0.144 & 4.030\\ 
         $-$ TD($\lambda$) & 0.040 & 0.278 & 4.098\\
         $-$ Exploration Policy & 0.063 & 0.346 & 4.160 \\
         \midrule
         VGS-IGNN & 0.023 & 0.111 & 4.044\\
         \bottomrule
    \end{tabular}
    }
    \vskip -0.2cm
\end{table}
}

{
\setlength{\tabcolsep}{0.5pt}
         
\begin{table*}[t]
    \centering
    \caption{Results of synthetic distribution experiments. The metrics are evaluated with 2,000 samples. We report the average and standard deviation over three random seeds. Metrics shown in \textbf{bold} denote the best mean performance, while \underline{underlined} entries indicate means are not statistically distinguishable from the best mean, using a one-sided Welch's t-test with a significance threshold of $p < 0.1$. For details, see \cref{appendix:metrics}
    and \cref{appendix:synthetic_details}.
    } 
    %\cref{appendix:metrics} and \cref{appendix:synthetic_details}
    \label{tab:main}
    % \scriptsize
    \resizebox{1\linewidth}{!}{
    \begin{tabular}{lccccccccc}
        \toprule
         Energy & \multicolumn{3}{c}{25GMM 
 ($D$=2)} &\multicolumn{3}{c}{Funnel  ($D$=10)} & \multicolumn{3}{c}{Manywell ($D$=32)}  \\
         \cmidrule(lr){1-1}\cmidrule(lr){2-4}\cmidrule(lr){5-7}\cmidrule(lr){8-10}
          Metric & $|\Delta \log Z_{r} |$ & $|\Delta \log Z_{f} |$ & $\mathcal{W}_{2}$ & $|\Delta \log Z_{r} |$ & $|\Delta \log Z_{f} |$ & $\mathcal{W}_{2}$ & $|\Delta \log Z_{r} |$ & $|\Delta \log Z_{f} |$ & $\mathcal{W}_{2}$ \\
         \midrule
         SMC  & 
         \multicolumn{2}{c}{0.345\tiny{$\pm$0.073}}& 5.207\tiny{$\pm$0.286} & \multicolumn{2}{c}{0.614\tiny{$\pm$0.100}} & \underline{31.495\tiny{$\pm$15.342}} & \multicolumn{2}{c}{30.170\tiny{$\pm$0.557}} & 5.882\tiny{$\pm$0.237} \\
         AFT & \multicolumn{2}{c}{0.092\tiny{$\pm$0.052}} & 4.930\tiny{$\pm$0.175} & \multicolumn{2}{c}{0.197\tiny{$\pm$0.122}}& 23.516\tiny{$\pm$5.167} & \multicolumn{2}{c}{13.692\tiny{$\pm$4.730}} & 8.905\tiny{$\pm$0.384} \\
         CRAFT & \multicolumn{2}{c}{0.131\tiny{$\pm$0.108}} & 4.938\tiny{$\pm$0.257} & \multicolumn{2}{c}{0.126\tiny{$\pm$0.059}} & 26.269\tiny{$\pm$1.960} & \multicolumn{2}{c}{13.076\tiny{$\pm$4.453}} & 9.244\tiny{$\pm$0.579} \\
         % \midrule
         % iDEM  & - &
         % - & 
         % - & 
         % - & 
         % - & 
         % - & 
         % - & 
         % - & 
         % - \\
         % DiKL  & \multicolumn{2}{c}{-} & 1.531\tiny{$\pm$0.165} & \multicolumn{2}{c}{-} & 31.721\tiny{$\pm$10.442} & \multicolumn{2}{c}{-} &  6.914\tiny{$\pm$0.046} \\
         \midrule
         PIS-LV  & 1.020\tiny{$\pm$0.006} & 0.248\tiny{$\pm$0.050} & 4.577\tiny{$\pm$0.039} & \underline{0.357\tiny{$\pm$0.373}} & 0.253\tiny{$\pm$0.037} & 22.567\tiny{$\pm$2.753} & 0.721\tiny{$\pm$0.058} & \textbf{0.134\tiny{$\pm$0.078}} & 5.462\tiny{$\pm$0.029}   \\
         DIS-LV  & 1.083\tiny{$\pm$0.004} &0.034\tiny{$\pm$0.018} & 4.656\tiny{$\pm$0.044} & \textbf{0.035\tiny{$\pm$0.019}} & 0.337\tiny{$\pm$0.014} & 20.630\tiny{$\pm$1.471} & 2.172\tiny{$\pm$0.302} &1.253\tiny{$\pm$0.642} & 6.144\tiny{$\pm$0.025}  \\
         DDS-LV  & 1.018\tiny{$\pm$0.002} & - & 4.678\tiny{$\pm$0.087} & 0.070\tiny{$\pm$0.017}  & - & 20.681\tiny{$\pm$3.627} & 2.281\tiny{$\pm$0.314} & - & 6.021\tiny{$\pm$0.097}  \\
         \midrule
         GFN-DB  & 0.559\tiny{$\pm$0.053} &
         0.071\tiny{$\pm$0.050} & 
         4.976\tiny{$\pm$0.159} & 
         0.552\tiny{$\pm$0.095} & 
         0.040\tiny{$\pm$0.017} & 
         28.070\tiny{$\pm$9.069} & 
         8.321\tiny{$\pm$5.800} & 
         7.057\tiny{$\pm$2.670} & 
         6.719\tiny{$\pm$0.531} \\
         GFN-TB  & 1.035\tiny{$\pm$0.015} & \underline{0.014\tiny{$\pm$0.011}} & 4.685\tiny{$\pm$0.055} & 0.284\tiny{$\pm$0.073} & \underline{0.016\tiny{$\pm$0.015}} &
         27.684\tiny{$\pm$9.120} & 2.726\tiny{$\pm$0.109} & \underline{0.198\tiny{$\pm$0.184}} & 6.161\tiny{$\pm$0.027} \\
         GFN-SubTB  & 0.016\tiny{$\pm$0.009} & \underline{0.013\tiny{$\pm$0.016}} & \underline{1.334\tiny{$\pm$0.090}} & 0.384\tiny{$\pm$0.039} & \underline{0.013\tiny{$\pm$0.010}} & 20.683\tiny{$\pm$2.514} & 2.660\tiny{$\pm$0.045} & 0.337\tiny{$\pm$0.051} & 6.092\tiny{$\pm$0.013}  \\
         GFN-TB-Imp  & 0.016\tiny{$\pm$0.009} & 0.018\tiny{$\pm$0.004} & \underline{1.186\tiny{$\pm$0.221}} & 0.154\tiny{$\pm$0.024} & 0.028\tiny{$\pm$0.016} & 23.497\tiny{$\pm$3.416} & \textbf{0.153\tiny{$\pm$0.051}} & \underline{0.243\tiny{$\pm$0.297}} & \textbf{5.330\tiny{$\pm$0.030}}  \\
         %\quad -Langevin &          \textbf{0.008\scriptsize{$\pm$0.004}} & \textbf{0.010\scriptsize{$\pm$0.002}} & 1.450\scriptsize{$\pm$0.069} & 0.293\scriptsize{$\pm$0.079} & 0.080\scriptsize{$\pm$0.044} & 25.337\scriptsize{$\pm$4.708} & 0.768\scriptsize{$\pm$0.458} & 1.215\scriptsize{$\pm$0.493} & 5.431\scriptsize{$\pm$0.029} \\ 
         \midrule 
        VGS (Ours) & \textbf{0.003\tiny{$\pm$0.005}} & \underline{0.022\tiny{$\pm$0.020}} & \underline{1.253\tiny{$\pm$0.107}} & 0.203\tiny{$\pm$0.082} & 0.040\tiny{$\pm$0.022} & 17.506\tiny{$\pm$0.769} & 2.427\tiny{$\pm$0.812} & 3.583\tiny{$\pm$2.775} & 5.666\tiny{$\pm$0.0.069}  \\
        \quad+Buffer & \underline{0.021\tiny{$\pm$0.016}} & \underline{0.021\tiny{$\pm$0.013}} & \textbf{1.181\tiny{$\pm$0.067}} & 0.168\tiny{$\pm$0.087} & 0.046\tiny{$\pm$0.031} &
        21.787\tiny{$\pm$4.134} & 1.360\tiny{$\pm$1.565} & 2.562\tiny{$\pm$0.354} & 5.474\tiny{$\pm$0.032} \\ 
        \quad+TD($\lambda$) & \underline{0.006\tiny{$\pm$0.001}} & \textbf{0.012\tiny{$\pm$0.004}} & \underline{1.214\tiny{$\pm$0.104}} & 0.208\tiny{$\pm$0.070} & \textbf{0.005\tiny{$\pm$0.001}} & \textbf{16.325\tiny{$\pm$0.797}} & 0.845\tiny{$\pm$0.050} & 0.517\tiny{$\pm$0.277} & 5.492\tiny{$\pm$0.012} \\
         \bottomrule
    \end{tabular}
    }
    \vskip -0.2cm
\end{table*}
}
\begin{figure}[t]
    \centering
    \includegraphics[width=1\linewidth]{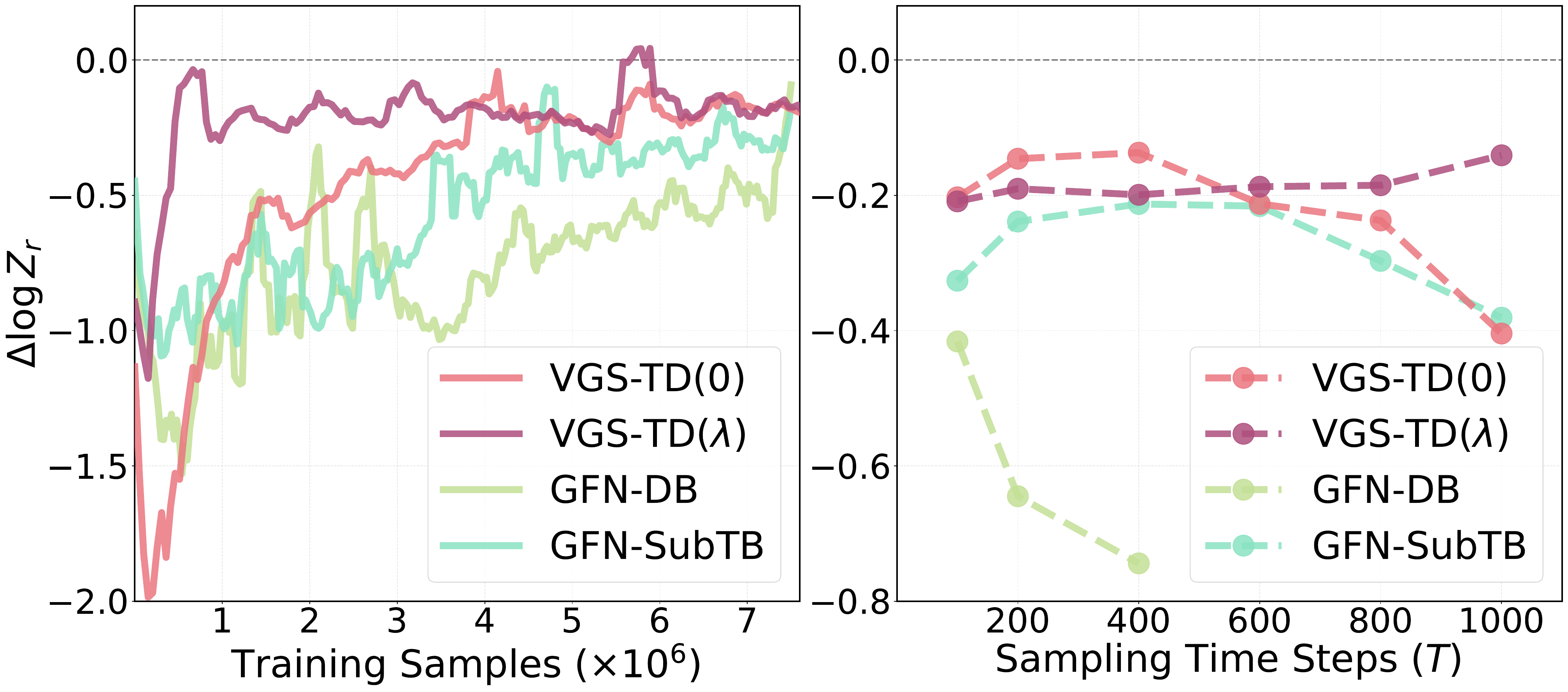}
    \vskip -0.2cm
    \caption{(\textbf{Left}) Evolution of $\Delta\log Z_r$ during training. (\textbf{Right}) Effect of varying $T$. VGS can reliably learn with a large number of time steps using TD($\lambda$). GFN-DB diverged when $T>400$. Experiments are conducted on Funnel.
    }
    \label{fig:td_lambda_analysis}    
    \vskip -0.2cm
\end{figure}

\textbf{Sampling Time.}\quad \cref{tab:sampling_time} reports the sampling times measured on LJ-55, where scalability poses a challenge. VGS-IMLP achieves the fastest per-step and total sampling times among all baselines, despite utilizing the largest number of parameters. Specifically, its per-step sampling is over $100\times$ faster than other models. In terms of total time, it is faster than the single-step generator DiKL \citep{he2024training} and approximately $1,000\times$ faster than iDEM \citep{akhound-sadegh2024iterated}, the best-performing baseline. Note that VGS uses the network's gradient during sampling, which incurs additional computational overhead. However, our results indicate that the efficiency of IMLP over equivariant networks outweighs this overhead. VGS-IGNN demonstrates a moderate sampling time; while slower than single-step generators, it remains faster than all other multi-step baselines.

% Among diffusion-based samplers, VGS with IMLP achieves the shortest sampling time, despite utilizing the largest number of parameters. In contrast, iDEM, PIS, aDDS are based on EGNN, which results in slower sampling speeds.
% Note that when the same network is used, VGS can be slower and require more memory during inference due to the overhead of value gradient computation. However, the efficiency of IMLP outweighs this overhead.
% VGS combined with invariant networks achieves faster per-step inference time than baselines employing equivariant networks, highlighting the contribution of our proposed architecture to efficient sampling. Moreover, VGS attains shorter total inference time than all other methods, except for DiKL \citep{he2024training} which operates with only a single time step. %This discrepancy primarily arises from the differences in the number of timesteps. Notably, when normalized by the number of timesteps, VGS demonstrates the fastest inference time among all methods. These results suggest that the use of our proposed invariant architecture contributes to the fast sampling of VGS.

\textbf{Importance of Invariance.}\quad
\cref{fig:lj13_histogram} visualizes the energy and interatomic distance histograms for LJ-13 from VGS. Three different neural network architectures, IGNN, IMLP, and conventional non-invariant MLP, are compared. 
The distributions of distances and energy align well with the test data only when the invariance is respected using IGNN and IMLP.
% Notably, IMLP achieves qualitatively   IGNN (\cref{tab:sampling_time}).
% For value function approximation, we employed three different networks: our proposed invariant architecture, IGNN, and a non-invariant network. 
% VGS demonstrates the ability to generate particle systems across different network architectures, provided invariance is ensured, highlighting the importance of invariant architectures in sampling n-body particle systems. While similar in performance, our proposed architecture is specifically designed for n-body particle systems and avoids the need for complex graph-structured data during training. Meanwhile, VGS combined with IGNN indicates its potential applicability to more general sampling problems where EGNNs \citep{satorras2021egnn} are conventionally employed.

\textbf{RL Techniques in VGS.}\quad We confirm the benefit of the RL techniques introduced in \cref{sec:exploration} by ablating each method (see \cref{tab:ablation}). Removing the off-policy exploration method brings the most significant degradation in performance metrics. Using TD($\lambda$) and the double value function techniques are also important in achieving the best performance. 
% Meanwhile, using IGNN instead of IMLP gives comparable performance.
% TD($\lambda$) facilitates long-term credit assignment in value function, resulting in faster and more stable training as further analyzed in \cref{sec:synthetic_exp}. Finally, using the mean value reduces fluctuations and stabilizes training; however, this effect is not fully reflected in performance due to our evaluation protocol. 

\subsection{Sampling from Synthetic Distributions}
\label{sec:synthetic_exp}

\textbf{Target Distributions.}\quad
We use three distributions as our sampling benchmarks: a 25-component Gaussian Mixture Model (25GMM), a funnel distribution, and a manywell distribution.

% and a MNIST distribution trained using NICE \cite{dinh2015nicenonlinearindependentcomponents}These benchmarks are chosen to demonstrate that our VGS produces results that are competitive with the baseline methods.
\textbf{Baselines.}\quad We compare VGS against four classes of baseline samplers on each target distribution: Sequential Monte Carlo–based methods (SMC \citep{moral2006smc}, AFT \citep{arbel2021annealedflowtransportmonte}, and CRAFT \citep{matthews2023continualrepeatedannealedflow}); SDE–based methods (PIS, DIS \citep{berner2022optimal}, and DDS with log-variance objective \citep{richter2024improved}); and GFlowNets (GFN-DB \citep{bengio2023gflownetfoundations}, GFN-TB \citep{lahlou2023theory}, GFN-SubTB \citep{zhang2023diffusion} and GFN-TB-Imp \citep{sendera2024improved}).

%denoising score matching–based methods (DiKL \citep{he2024training} and iDEM \citep{akhound-sadegh2024iterated});
 
\textbf{Performance Metrics.}\quad
 The comparisons are made based on three metrics: the bias in reverse importance-weighted estimation of log partition function ($|\Delta \log Z_{r}| \downarrow$), the bias in forward importance-weighted estimation of log partition function ($|\Delta \log Z_{f}| \downarrow$) \citep{blessing2024elboslargescaleevaluationvariational} and the Wasserstein-2 distance ($\mathcal{W}_{2} \downarrow$).

% \textbf{VGS Implementation.}\quad 
%  The trainable value function of the VGS sampler $V^t_\phi:\mathbb{R}^d \times \mathbb{Z}^{+} \rightarrow \mathbb{R}$ can be trained using any neural network architecture that incorporates time step embedding. We use a Fourier MLP architecture, an MLP with sinusoidal time step embedding.

\textbf{Results.}\quad
We evaluate VGS trained with TD(0), TD($\lambda$), and sample buffers, with the results summarized in \cref{tab:main}. We set $T=50$ for 25GMM, and $T=100$ for Funnel and Manywell. For all TD($\lambda$) experiments, we use $\lambda = 0.9$. 

VGS-TD(0) and VGS-TD($\lambda$) achieve performance comparable to baseline models that do not employ additional exploration strategies (e.g., PIS-LV, GFN-TB). When a sample buffer is used, VGS shows results that are on par with baselines incorporating exploration mechanisms (e.g., GFN-TB-Imp).

\textbf{Benefit of TD($\lambda$).}\quad 
%We focus on comparing our algorithm directly with the GFN-TB method and its variants, since each VGS variant has a one-to-one correspondence with a GFN variant via an RL analogue.
%The detailed-balance objective \cite{bengio2021gflownet} can be computed using two consecutive time steps from a sample, resembling the temporal-difference objective in RL. In our framework, it is exactly equivalent to VGS-TD(0). On the other hand, the GFN trajectory-balance objective \cite{lahlou2023theory} requires rolling out an entire trajectory, which is analogous to Monte Carlo sampling and is equivalent to VGS-TD(1).
%Inspired from TD($\lambda$), the sub-trajectory balance objective \cite{madan2023learning} mixes detailed balance and trajectory balance objective by leveraging observed sub-trajectories. This is equivalent to VGS-TD($\lambda$). Finally, just as off-policy exploration methods have been used to improve the training of GFN \cite{sendera2024improved}, VGS-Exploration incorporates off-policy exploration for further improvements. 
\cref{fig:td_lambda_analysis} compares TD(0) and TD($\lambda$) applied to VGS. For comparison, we also present GFN-DB and GFN-SubTB. Specifically, TD(0) and GFN-DB are aligned in that they both minimize single-step differences, whereas TD($\lambda$) and GFN-SubTB share the common feature of operating over subtrajectories. VGS-TD(0) exhibits more stable training dynamics than GFN-DB, as expected from the use of target networks. TD($\lambda$) demonstrates improved performance at larger $T$, addressing the limitations of TD(0) in long-term credit assignment.
% VGS-TD($\lambda$), designed to address the limitations of TD(0) in long-term credit assignment, demonstrates improved performance at larger timesteps, consistent with its design intent. Moreover, VGS-TD($\lambda$) converges faster than TD(0), providing an additional advantage.

\textbf{Training with Buffer.}\quad 
We utilize sampler buffer as introduced in \cref{sec:exploration}.
To quantify the impact of the buffer, we compare VGS with and without the buffer in \cref{tab:main}. We adopt the buffer design and MCMC exploration strategy from \cite{sendera2024improved}.
The performance gain from the buffer is not always consistent, possibly due to sensitivity to the hyperparameters of the MCMC exploration.
% Although not always, using the buffer often improves performance compared to arbitrary TD($\lambda$) setting, clearly to TD(0). This highlights the potential applicability and compatibility of other RL techniques beyond those we implemented when external samples are provided.

%The full result and experimental details are in \cref{appendix:anomaly}.

% The EBM trained with VGS outperforms an EBM trained with a sophisticated MCMC scheme

% which contains 224$\times$224 RGB images of 15 object categories. We follow the multi-class problem setup proposed by \cite{you22uniad}. The training dataset contains normal object images from 15 categories without any labels. The test set consists of both normal and defective object images, each provided with an anomaly label and a mask indicating the defect location. The goal is to detect and localize anomalies, with performance measured by AUC computed per object category. This setting is challenging because the energy function should reflect the multi-modal data distribution.
% Following the preprocessing protocol in \cite{you22uniad, yoon2023energybased}, each image is transformed into a 272$\times$14$\times$14 vector using a pre-trained EfficientNet-b4 \cite{tan2019efficientnet}. VGS is conducted in a 272-dimensional space, treating each spatial coordinate independently. With the trained energy function, we can evaluate the energy value of 14x14 spatial features and use max pooling and bilinear interpolation for anomaly detection and localization, respectively.

\section{Related Work}

VGS builds on diffusion-based samplers, where the learned trajectory corresponds to the reversed diffusion process. Following diffusion generative models \citep{ho2020ddpm,song2021scorebased}, samplers have been developed to model the stochastic bridge between initial and target distributions. Some train neural SDEs \citep{zhang2022path,berner2022optimal,vargas2023denoising,richter2024improved,havens2025adjoint}, while others model PDEs \citep{shi2024diffusion,sun2024dynamical}. A central challenge is estimating the score of diffused densities, since denoising score matching is inapplicable without training data. Alternatives rely on various mathematical formulations \citep{akhound-sadegh2024iterated,phillips2024particle,huang2024reverse,mcdonald2022proposal,wang2024energy,chen2024sequential,he2024training,vargas2024transport}. GFlowNets view sampling as a sequential decision-making problem \citep{bengio2021gflownet,bengio2023gflownetfoundations,malkin2022trajectorybalance,lahlou2023theory, madan2023learning}.
Similar to GFlowNets, VGS sidesteps score estimation via value-based RL, estimating the value function corresponding to the energy of the target-reference density ratio. The idea of leveraging value functions and TD learning in diffusion was explored in \citet{yoon2024maximum}.\looseness=-1

\section{Conclusion}
\label{sec:conlusion}

This paper introduces the Value Gradient Sampler (VGS), a diffusion sampler defined by value functions and trained via RL. VGS is particularly effective and efficient when sampling from target densities that exhibit invariance symmetries.

\textbf{Limitations.}\quad 
First, VGS needs to compute gradients during sampling, which imposes overhead in both computation and memory. However, in particle system applications, this drawback is outweighed by the ability to use more efficient network architectures.
Second, we were not able to experiment with a recent large-scale conformer sampling benchmark \citep{havens2025adjoint} due to computational constraints. 
Third, no theoretical guarantee is provided for VGS. 
We left larger-scale applications and rigorous mathematical analysis as future work. \looseness=-1
% First, VGS does not provide an exact likelihood estimation for the generated samples, which prevents the direct use of techniques such as importance sampling, which relies on precise likelihood values. Second, the error of VGS is not mathematically quantified. We leave rigorous mathematical analysis as future work. 

\subsubsection*{Acknowledgements}
H. Hwang, H. Jeong, C. Park, S. Kweon, and F. Park were supported in part by IITP-MSIT under Grants 2022-220480 and RS-2022-II220480 (Training and Inference Methods for Goal Oriented AI Agents), and Grant RS-2024-00436680 (Collaborative Research Projects with Microsoft Research); KIAT under Grant P0020536 (HRD Program for Industrial Innovation); MOTIR under Grants RS-2025-25462891 and RS-2025-25460896; Hyundai Motor Company and Kia; Microsoft Research Asia; and SNU-IPAI, SNU-AIIS, SNU-IAMD, the SNU BK21+ Program in Mechanical Engineering, and the SNU Institute for Engineering Research.
S. Yoon was supported by the Institute of Information \& Communications Technology Planning \& Evaluation (IITP) grant funded by the Korea government (MSIT) (No.\,RS-2020-II201336, Artificial Intelligence Graduate School Program (UNIST) and No.\,RS-2025-25442824, AI Star Fellowship Program (Ulsan National Institute of Science and Technology)), the National Research Foundation of Korea (NRF) grant funded by the Korea government (MSIT) (No. RS-2024-00408003), and the Center for Advanced Computation at Korea Institute for Advanced Study.

% All acknowledgments go at the end of the paper, including thanks to reviewers who gave useful comments, to colleagues who contributed to the ideas, and to funding agencies and corporate sponsors that provided financial support. 
% To preserve the anonymity, please include acknowledgments \emph{only} in the camera-ready papers. The acknowledgements do not count against the 9-page page limit in the camera-ready.

% \subsubsection*{References}
\begin{small}
\bibliography{4_ref}
\end{small}

%%%%%%%%%%%%%%%%%%%%%%%%%%%%%%%%%%%%%%%%%%%%%%%%%%%%%%%%%%%%
\section*{Checklist}

% %%% BEGIN INSTRUCTIONS %%%
% The checklist follows the references. For each question, choose your answer from the three possible options: Yes, No, Not Applicable.  You are encouraged to include a justification to your answer, either by referencing the appropriate section of your paper or providing a brief inline description (1-2 sentences). 
% Please do not modify the questions.  Note that the Checklist section does not count towards the page limit. Not including the checklist in the first submission won't result in desk rejection, although in such case we will ask you to upload it during the author response period and include it in camera ready (if accepted).

% \textbf{In your paper, please delete this instructions block and only keep the Checklist section heading above along with the questions/answers below.}
% %%% END INSTRUCTIONS %%%

\begin{enumerate}

  \item For all models and algorithms presented, check if you include:
  \begin{enumerate}
    \item A clear description of the mathematical setting, assumptions, algorithm, and/or model. Yes.
    \item An analysis of the properties and complexity (time, space, sample size) of any algorithm. Yes.
    \item (Optional) Anonymized source code, with specification of all dependencies, including external libraries. No, but the code will be open-sourced once the manuscript is published.
  \end{enumerate}

  \item For any theoretical claim, check if you include:
  \begin{enumerate}
    \item Statements of the full set of assumptions of all theoretical results. Yes.
    \item Complete proofs of all theoretical results. Yes.
    \item Clear explanations of any assumptions. Yes.     
  \end{enumerate}

  \item For all figures and tables that present empirical results, check if you include:
  \begin{enumerate}
    \item The code, data, and instructions needed to reproduce the main experimental results (either in the supplemental material or as a URL). Yes.
    \item All the training details (e.g., data splits, hyperparameters, how they were chosen). Yes.
    \item A clear definition of the specific measure or statistics and error bars (e.g., with respect to the random seed after running experiments multiple times). Yes.
    \item A description of the computing infrastructure used. (e.g., type of GPUs, internal cluster, or cloud provider). Yes.
  \end{enumerate}

  \item If you are using existing assets (e.g., code, data, models) or curating/releasing new assets, check if you include:
  \begin{enumerate}
    \item Citations of the creator if your work uses existing assets. Yes.
    \item The license information of the assets, if applicable. Not Applicable.
    \item New assets either in the supplemental material or as a URL, if applicable. Not Applicable.
    \item Information about consent from data providers/curators. Not Applicable.
    \item Discussion of sensible content if applicable, e.g., personally identifiable information or offensive content. Not Applicable.
  \end{enumerate}

  \item If you used crowdsourcing or conducted research with human subjects, check if you include:
  \begin{enumerate}
    \item The full text of instructions given to participants and screenshots. Not Applicable.
    \item Descriptions of potential participant risks, with links to Institutional Review Board (IRB) approvals if applicable. Not Applicable.
    \item The estimated hourly wage paid to participants and the total amount spent on participant compensation. Not Applicable.
  \end{enumerate}

\end{enumerate}

\clearpage
\appendix
\thispagestyle{empty}

% Supplementary material: To improve readability, you must use a single-column format for the supplementary material.
\onecolumn
\aistatstitle{Supplementary Materials for Value Gradient Sampler: Learning Invariant Value Functions for Equivariant Diffusion Sampling}

%%%%%%%%%%%%%%%%%%%%%%%%%%%%%%%%%%%%%%%%%%%%%%%%%%%%%%%%%%%%%%%%%%%%%%%%%%%%%%
%%%%%%%%%%%%%%%%%%%%%%%%%%%%%%%%%%%%%%%%%%%%%%%%%%%%%%%%%%%%%%%%%%%%%%%%%%%%%%
%APPENDIX
%%%%%%%%%%%%%%%%%%%%%%%%%%%%%%%%%%%%%%%%%%%%%%%%%%%%%%%%%%%%%%%%%%%%%%%%%%%%%%
%%%%%%%%%%%%%%%%%%%%%%%%%%%%%%%%%%%%%%%%%%%%%%%%%%%%%%%%%%%%%%%%%%%%%%%%%%%%%%
% \newpage
% \appendix

\section{Proofs and Derivations}
\label{sec:proof}

\subsection{Derivation of Optimal control objective}
\label{appendix:objective_dervation}
We start from the joint KL-divergence minimization objective:
\begin{align}
    \min_{\pi\in\Pi}\text{KL}(\pi(\bx_{0:T})\Vert \tilde{q}(\bx_{0:T})) =\min_{\pi\in\Pi} \mathbb{E}_{\pi(\bx_{0:T})} \left[
\log{\frac{\pi(\bx_{0:T})}{\tilde{q}(\bx_{0:T})}}
\right]
\end{align}
Substituting our choice of the joint target distribution $\tilde{q}(\bx_{0:T}) = \tilde{\pi}(\bx_{0:T})q(\bx_T)/\tilde{\pi}(\bx_T)$ and using \cref{eq:boltzmann}, and dropping constant terms, we obtain
\begin{align}\label{eq:intermid_obj}
    \min_{\pi\in\Pi} \mathbb{E}_{\pi(\bx_{0:T})} \left[\log{\frac{\pi(\bx_0)}{\tilde{\pi}(\bx_0)}}+
\sum_{t=0}^{T-1} \log{\frac{\pi(\bx_{t+1}|\bx_t
)}{\tilde{\pi}(\bx_{t+1}|\bx_t)}} + E(\bx_T) + \log{\tilde{\pi}(\bx_T)}
\right]
\end{align}
From our construction of $\pi$ (\cref{eq:pls}) and the reference distribution $\tilde{\pi}$, we have $\pi(\bx_0)/\tilde{\pi}(\bx_0)=1$, and
\begin{align}\label{eq:obj_simplification}
    \mathbb{E}_{\pi_{\phi}(\bx_{t+1}|\bx_t)}\left[\log{\frac{\pi(\bx_{t+1}|\bx_t)}{\tilde{\pi}(\bx_{t+1}|\bx_t)}}\bigg| \bx_t\right] = \mathbb{E}_{\epsilon_t\sim \mathcal{N}(0, I)}\left[-\frac{\Vert \sigma_t\epsilon_t\Vert^2}{2\sigma_t^2} + \frac{\Vert \mu^t(\bx_t) + \sigma_t\epsilon_t\Vert^2}{2\sigma_t^2}\right] = \frac{\Vert \mu^t(\bx_t)\Vert^2}{2\sigma_t^2},
\end{align}
Therefore, by the tower property, \cref{eq:intermid_obj} can be rewritten as
\begin{align}
    \min_{\pi\in\Pi} \mathbb{E}_{\pi(\bx_{0:T})} \left[
\sum_{t=0}^{T-1} \frac{\Vert \mu^t(\bx_t)\Vert^2}{2\sigma_t^2} + \tilde{E}(\bx_T)
\right],
\end{align}
where $\tilde{E}(\bx_T) = E(\bx_T) + \log{\tilde{\pi}(\bx_T)}$.

\subsection{Derivation of the Optimal Value Function}
\label{appendix:opt_val_proof}
For the proof, using \cref{eq:obj_simplification} we rewrite the value function back to its general form:
\begin{align}
V^{t}_\pi(\bx_t)
&=\mathbb{E}_{\pi(\cdot| \bx_t)}\left[
\sum_{i=0}^{T-t-1} \frac{\Vert\mu^{t+i}(\bx_{t+i})\Vert^2}{2\sigma_{t+i}^2} + \tilde{E}(\bx_T) \bigg| \bx_t \right] \\
&= \mathbb{E}_{\pi(\cdot| \bx_t)}\left[
\sum_{i=0}^{T-t-1} \log\frac{\pi(\bx_{t+1}| \bx_t)}{\tilde{\pi}(\bx_{t+1}| \bx_t)} + \tilde{E}(\bx_T) \bigg| \bx_t \right]\\
&= \mathbb{E}_{\pi(\cdot| \bx_t)}\left[\log\frac{\pi(\bx_{t+1:T}| \bx_t)}{\tilde{\pi}(\bx_{t+1:T}| \bx_t)}+ \log\frac{\tilde{\pi}(\bx_T)}{q(\bx_T)} - \log Z\right]\\
&= \mathbb{E}_{\pi(\cdot| \bx_t)}\left[\log\frac{\pi(\bx_{t+1:T}| \bx_t)}{\tilde{q}(\bx_{t+1:T}| \bx_t)}+ \log\frac{\tilde{\pi}(\bx_t)}{\tilde{q}(\bx_t)} - \log Z\right].
\end{align}
The last equality follows from Bayes’ rule applied to our choice of the joint target distribution, $\tilde{q}(\bx_{t:T})=\tilde{\pi}(\bx_{t:T})q(\bx_T)/\tilde{\pi}(\bx_T)$.

Now, by the definition of the optimal value function,
\begin{align}
V_*^t(\bx_t)
&= \min_{\pi(\cdot| \bx_t)\in \Pi} V_{\pi}^t(\bx_t) \\
&= \log\frac{\tilde{\pi}(\bx_t)}{\tilde{q}(\bx_t)} - \log Z
+ \min_{\pi(\cdot| \bx_t)\in \Pi}\mathbb{E}_{\pi(\cdot|\bx_t)}
\left[\log\frac{\pi(\bx_{t+1:T}| \bx_t)}{\tilde{q}(\bx_{t+1:T}| \bx_t)}\right]\\
&= \log\frac{\tilde{\pi}(\bx_t)}{\tilde{q}(\bx_t)} - \log Z
+ \min_{\pi(\cdot| \bx_t)\in \Pi}
\text{KL}\big(\pi(\bx_{t+1:T}| \bx_t) \Vert \tilde{q}(\bx_{t+1:T}| \bx_t)\big)\\
&= \log\frac{\tilde{\pi}(\bx_t)}{\tilde{q}(\bx_t)} - \log Z.
\end{align}
The last step uses the assumption that $\tilde{q}(\bx_{t+1:T}| \bx_t)\in \Pi$, so the minimum KL divergence is attained at zero.

Exponentiating both sides and rearranging terms gives the desired relation:
\begin{align}
\frac{\tilde{q}(\bx_t)}{\tilde{\pi}(\bx_t)} = \frac{1}{Z}\exp{(-V_*^t(\bx_t))}.
\end{align}

\subsection{Derivation of VGS}\label{appendix:second_order}
In this section, we present a detailed derivation of VGS. To obtain an approximate solution to the optimization problem in \cref{eq:vgs-obj}, we apply a first-order Taylor expansion of $V_{\pi}^{t+1}$ around $\alpha_t \bx_t$:
\begin{align}
\mathbb{E}_{\pi(\bx_{t+1}|\bx_t)} \left[ \frac{\Vert\mu^t(\bx_t)\Vert^2}{2\sigma_t^2} +V^{t+1}_{\pi}(\bx_{t+1})\bigg| \bx_{t}\right] &= \frac{\Vert\mu^t(\bx_t)\Vert^2}{2\sigma_t^2} + \mathbb{E}_{\epsilon_t\sim \mathcal{N}(0, I)}\left[V_{\pi}^{t+1}(\alpha_t\bx_t + \mu^t(\bx_t)+\sigma_t\epsilon_t)\right]\\
&\approx \frac{\Vert\mu^t(\bx_t)\Vert^2}{2\sigma_t^2} + \mathbb{E}_{\epsilon_t\sim \mathcal{N}(0, I)}\left[V_{\pi}^{t+1}(\alpha_t\bx_t) + (\mu^t(\bx_t)+\sigma_t\epsilon_t)^T\nabla_{\alpha_t \bx_t}V_{\pi}^{t+1}(\alpha_t\bx_t) \right] \label{eq:first_order_approx}\\
&=\frac{\Vert\mu^t(\bx_t)\Vert^2}{2\sigma_t^2} + V_{\pi}^{t+1}(\alpha_t\bx_t) + \mu^t(\bx_t)^T\nabla_{\alpha_t \bx_t}V_{\pi}^{t+1}(\alpha_t\bx_t).
\end{align}
Substituting this approximation into \cref{eq:vgs-obj} and dropping constant terms, we obtain the following surrogate problem:
\begin{align}
\min_{\mu^t(\bx_t)} \frac{\Vert\mu^t(\bx_t)\Vert^2}{2\sigma_t^2} + \mu^t(\bx_t)^T\nabla_{\alpha_t \bx_t}V_{\pi}^{t+1}(\alpha_t\bx_t).
\end{align}
Setting the derivative with respect to $\mu^t(\bx_t)$ to zero yields the closed-form optimal drift:
\begin{align}
\mu^t(\bx_t) = -\sigma_t^2\nabla_{\alpha_{t}\bx_{t}} V^{t+1}_{\pi}(\alpha_{t}\bx_t).\label{eq:first_order_mu}
\end{align}
We use this first-order approximation throughout the main text and experiments. However, a second-order Taylor expansion in \cref{eq:first_order_approx} also admits an analytic solution, which we include here for completeness. In that case, the optimal drift is
\begin{align}
\mu^t(\bx_t) = - \left(I + \sigma_t^2 \nabla^2_{\alpha_t\bx_t}V_{\pi}^{t+1}(\alpha_t\bx_t) \right)^{-1} \sigma_t^2 \nabla_{\alpha_t\bx_t}V_{\pi}^{t+1}(\alpha_t\bx_t).
\end{align}

\subsection{Off-policy TD(\texorpdfstring{$\lambda$}{lambda})}\label{appendix:off_td_lambda}
Recall that the TD($\lambda$) target for on-policy samples $\bx_{0:T}\sim \pi_{\phi^-}(\bx_{0:T})$ is given by \cref{eq:td_lambda_target}. Now suppose a trajectory is collected under a different policy $\pi_{\text{expl}}$, i.e., $\bx_{0:T}\sim \pi_{\text{expl}}(\bx_{0:T})$. Using importance sampling, we can construct an unbiased estimator of the on-policy TD($\lambda$) target's expectated value $\mathbb{E}_{\pi_{\phi^-}(\cdot | \bx_t)}\left[\hat{V}_{\text{TD}(\lambda)}^{t}(\bx_{t:T})\right]$ based on samples from $\pi_{\text{expl}}$. Building on \cref{eq:td_lambda_target} and applying the tower property, we have
\begin{align}
    &\mathbb{E}_{\pi_{\phi^-}(\bx_{t+1:T}|\bx_t)}[\hat{V}_{\text{TD}(\lambda)}^t(\bx_{t:T})] \\
    &= \mathbb{E}_{\pi_{\phi^-}(\bx_{t+1:T}|\bx_t)}\left[V_{\phi^-}^t(\bx_t) + \sum_{i=0}^{T-t-1} \lambda^i \delta_{t+i}(\bx_{t+i}, \bx_{t+i+1})\right] \\
    &= V_{\phi^-}^t(\bx_t) + \mathbb{E}_{\pi_{\phi^-}(\bx_{t+1:T}|\bx_t)}\left[\sum_{i=0}^{T-t-1} \lambda^i \mathbb{E}_{\pi_{\phi^-}(\bx_{t+i+1}|\bx_{t+i})}\left[\delta_{t+i}(\bx_{t+i}, \bx_{t+i+1})\right]\right] \\
    &= V_{\phi^-}^t(\bx_t) + \mathbb{E}_{\pi_{\text{expl}}(\bx_{t+1:T}|\bx_t)}\left[\sum_{i=0}^{T-t-1} \lambda^i \frac{\pi_{\phi^-}(\bx_{t+i}|\bx_t)}{\pi_{\text{expl}}(\bx_{t+i}|\bx_t)}\mathbb{E}_{\pi_{\phi^-}(\bx_{t+i+1}|\bx_{t+i})}\left[\delta_{t+i}(\bx_{t+i}, \bx_{t+i+1})\right]\right] \\
    &= V_{\phi^-}^t(\bx_t) + \mathbb{E}_{\pi_{\text{expl}}(\bx_{t+1:T}|\bx_t)}\left[\sum_{i=0}^{T-t-1}  \prod_{j=0}^{i-1}\left(\lambda\frac{\pi_{\phi^-}(\bx_{t+j+1}|\bx_{t+j})}{\pi_{\text{expl}}(\bx_{t+j+1}|\bx_{t+j})}\right) \mathbb{E}_{\pi_{\phi^-}(\bx_{t+i+1}|\bx_{t+i})}\left[\delta_{t+i}(\bx_{t+i}, \bx_{t+i+1})\right]\right].
\end{align}
Thus, given off-policy samples \(\bx_{0:T} \sim \pi_{\text{expl}}(\bx_{0:T})\), the importance-sampling–based estimate of the TD($\lambda$) target can be computed as
\begin{align}
    \hat{V}_{\text{off-TD}(\lambda)}^t(\bx_{t:T}) = V_{\phi^-}^t(\bx_t) + \sum_{i=0}^{T-t-1} \left(\prod_{j=0}^{i-1} \lambda \frac{\pi_{\phi^-}(\bx_{t+j+1}|\bx_{t+j})}{\pi_{\text{expl}}(\bx_{t+j+1}|\bx_{t+j})} \right) \delta_{t+i}(\bx_{t+i}, \bx'_{t+i+1}), \label{eq:td_target_net}
\end{align}
where $\delta_{t+i}(\bx_{t+i}, \bx'_{t+i+1})$ is the TD error computed with a freshly sampled $\bx'_{t+i+1} \sim \pi_{\phi^-}(\cdot|\bx_{t+i})$, and is therefore an unbiased estimator of $\mathbb{E}_{\pi_{\phi^-}(\bx_{t+i+1}|\bx_{t+i})}\left[\delta_{t+i}(\bx_{t+i}, \bx_{t+i+1})\right]$.
We used $\bx'_{t+i+1}$ to distinguish it from the off-policy trajectory sample $\bx_{t+i+1}$.
The targets $\hat{V}_{\text{off-TD}(\lambda)}^t(\bx_{t:T})$ for $t = 0, \dots, T-1$ can be efficiently computed in a recursive manner using the tuples $\mathcal{D} = \{(\bx_t, \mu_{\phi^-}^t(\bx_t))\}_{t=0}^{T}$ stored during sampling.  
The full procedure is described in \cref{alg:off_td_lambda}.

\subsection{Details on the Invariance of the Value Function}\label{appendix:thm4.1}
\textbf{Proposition 4.1} (Invariance of $V^t_{\pi}$ and $V_*^t$.) \textit{    Assume that the energy function is $\mathrm{O}(m)\times \mathrm{S}_n$–invariant as follows:
\begin{align}
    E((R, \sigma) \circ \bx) =  E(\bx) \quad \forall \bx\in \mathcal{X}, (R, \sigma)\in \mathrm{O}(m)\times \mathrm{S}_n. \nonumber
\end{align}
then both the value function of VGS and the optimal value function preserves $\mathrm{O}(m)\times \mathrm{S}_n$-invariance:
\begin{align}
    V^{t}_{\pi}((R, \sigma) \circ \bx_t) =  V^t_{\pi}(\bx_t), \quad 
    V^{t}_{*}((R, \sigma) \circ \bx_t) =  V^t_{*}(\bx_t), \nonumber
\end{align}
for all $\bx_t\in \mathcal{X}$, $(R, \sigma)\in \mathrm{O}(m)\times \mathrm{S}_n$.
}
\begin{proof}
In this proof, we omit the subscripts of $\nabla$ and write $\nabla V(\bx)$ in place of $\nabla_{\bx} V(\bx)$ for brevity.

\textbf{Invariance of $V^t_{\pi}$.}\quad Substituting $\mu_t(\bx_t) = -\sigma^2_t\nabla V^{t+1}_{\pi}(\alpha_t\bx_t)$ to \cref{eq:vgs-obj}, we obtain the recurrence relation for the value function of VGS
\begin{align}
    V^{t}_{\pi}(\bx_t) &= \frac{\sigma_t^2 \Vert \nabla V_{\pi}^{t+1}(\alpha_t\bx_t)\Vert^2}{2} + \mathbb{E}_{\pi(\bx_{t+1}|\bx_t)}\left[V_{\pi}^{t+1}(\bx_{t+1})\right]\\
    &=\frac{\sigma_t^2 \Vert \nabla V_{\pi}^{t+1}(\alpha_t\bx_t)\Vert^2}{2} + \mathbb{E}_{\epsilon_t\sim \mathcal{N}(0, I) }\left[V_{\pi}^{t+1}(\alpha_t\bx_t -\sigma_t^2\nabla V^{t+1}_{\pi}(\alpha_t\bx_t)+\sigma_t\epsilon_t)\right]. \label{eq:recurrence_v}
\end{align}
Assume $V_{\pi}^{t+1}$ is $\mathrm{O}(m)\times \mathrm{S}_n$–invariant. We show that $V_{\pi}^{t}$ inherits this invariance. First, the action $\circ$ acts orthogonally on $\mathcal{X}$: for all $\bx,\by\in\mathcal{X}$, $(R, \sigma)\in \mathrm{O}(m)\times \mathrm{S}_n$, 
\begin{align}
    \langle (R,\sigma) \circ \bx, (R,\sigma) \circ \by \rangle = \langle [Rx_{\sigma(1)}, ..., Rx_{\sigma(n)}],[Ry_{\sigma(1)}, ..., Ry_{\sigma(n)}] \rangle = \sum_{i=1}^{n}x_{\sigma(i)}^TR^TRy_{\sigma(i)} = \sum_{i=1}^{n}x^T_iy_i = \langle \bx, \by \rangle. \label{eq:invariance_orthogonal}
\end{align}
Hence, the invariance of $V_{\pi}^{t+1}$ implies the equivariance of its gradient under $\circ$:
$\nabla V_{\pi}^{t+1}\big((R,\sigma)\circ \bx\big)=(R,\sigma)\circ \nabla V_{\pi}^{t+1}(\bx)$ \citep[Lemma~2]{papamakarios2021normalizing}. Using this property and the recurrence relation in \cref{eq:recurrence_v}, we have for all $\bx_t\in \mathcal{X}$ and $(R, \sigma)\in \mathrm{O}(m)\times S_n$,
\begin{align}
    V^t_{\pi}((R, \sigma) \circ \bx_t) &= \frac{\sigma_t^2\Vert \nabla V_{\pi}^{t+1}((R, \sigma) \circ \alpha_t\bx_t)\Vert^2}{2} +\mathbb{E}_{\epsilon_t\sim \mathcal{N}(0, I)}\left[V_{\pi}^{t+1}((R, \sigma) \circ \alpha_t\bx_t - \sigma_t^2\nabla V_{\pi}^{t+1}((R, \sigma) \circ \alpha_t\bx_t)+\sigma_t\epsilon_t) \right]\\
    &= \frac{\sigma_t^2\Vert (R, \sigma) \circ \nabla V_{\pi}^{t+1}(\alpha_t\bx_t)\Vert^2}{2} +\mathbb{E}_{\epsilon_t\sim \mathcal{N}(0, I)}\left[V_{\pi}^{t+1}((R, \sigma) \circ( \alpha_t\bx_t - \sigma_t^2\nabla V_{\pi}^{t+1}(\alpha_t\bx_t)+\sigma_t(R, \sigma)^{-1} \circ \epsilon_t) \right]\\
    &= \frac{\sigma_t^2\Vert \nabla V_{\pi}^{t+1}(\alpha_t\bx_t)\Vert^2}{2} +\mathbb{E}_{ (R, \sigma)^{-1} \circ \epsilon_t \sim \mathcal{N}(0,I)}\left[V_{\pi}^{t+1}((R, \sigma) \circ (\alpha_t\bx_t - \sigma_t^2\nabla V_{\pi}^{t+1}(\alpha_t\bx_t)+\sigma_t(R, \sigma)^{-1} \circ \epsilon_t)) \right]
    \\
    &= \frac{\sigma_t^2\Vert \nabla V_{\pi}^{t+1}(\alpha_t\bx_t)\Vert^2}{2} +\mathbb{E}_{ \epsilon'_t \sim \mathcal{N}(0,I)}\left[V_{\pi}^{t+1}(\alpha_t\bx_t - \sigma_t^2\nabla V_{\pi}^{t+1}(\alpha_t\bx_t)+\sigma_t\epsilon'_t) \right]\\
    &= V_{\pi}^t(\bx_t).
\end{align}
The second equality uses gradient equivariance. The third uses that $\mathcal{N}(0,I)$ is invariant under rotations, reflections, and permutations, so $(R,\sigma)^{-1}\circ \epsilon_t\sim \mathcal{N}(0,I)$. The fourth equality uses the assumed invariance of $V_{\pi}^{t+1}$.

Finally, we verify the terminal case $t=T$. By construction of the reference process ($\mu=0$ in \cref{eq:pls}), $\tilde{\pi}(\bx_T)$ is an isotropic zero-mean Gaussian. Consequently, $V_{\pi}^{T}(\bx_T)=\tilde{E}(\bx_T)=E(\bx_T)+\log \tilde{\pi}(\bx_T)$ is $\mathrm{O}(m)\times \mathrm{S}_n$–invariant whenever $E(\bx_T)$ is invariant. The result then follows by backward induction on $t$.

\textbf{Invariance of $V^t_*$.} \quad By \cref{eq:opt_val}, it suffices to show that $\tilde{q}(\bx_t)$ inherits the $\mathrm{O}(m)\times \mathrm{S}_n$-invariance of $E(\bx_T)$, since $\tilde{\pi}(\bx_t)$ is an isotropic zero-mean Gaussian and therefore invariant. If $E(\bx_T)$ is $\mathrm{O}(m)\times \mathrm{S}_n$-invariant, then by \cref{eq:boltzmann} the density $q(\bx_T)$ is also $\mathrm{O}(m)\times \mathrm{S}_n$-invariant. Moreover, $\tilde{\pi}(\bx_{t+1}| \bx_t)$ is invariant in the sense that for all $\bx_t, \bx_{t+1}\in \mathcal{X}$ and $(R,\sigma)\in \mathrm{O}(m)\times \mathrm{S}_n$, we have
\begin{align}
    \tilde{\pi}((R, \sigma)\circ \bx_{t+1}|(R, \sigma)\circ \bx_t) &= \frac{1}{(2\pi\sigma_t^2)^{D/2}}\exp{-\frac{\Vert (R, \sigma) \circ \bx_{t+1} - (R, \sigma)\circ \alpha_t\bx_t \Vert^2}{2\sigma_t^2}}\\
    &=\frac{1}{(2\pi\sigma_t^2)^{D/2}}\exp{-\frac{\Vert (R, \sigma) \circ (\bx_{t+1} - \alpha_t\bx_t) \Vert^2}{2\sigma_t^2}}\\
    &= \frac{1}{(2\pi\sigma_t^2)^{D/2}}\exp{-\frac{\Vert\bx_{t+1} - \alpha_t\bx_t \Vert^2}{2\sigma_t^2}}\\
    &= \tilde{\pi}(\bx_{t+1}|\bx_t).
\end{align}
The third equality follows from the orthogonality of the action~$\circ$, as established in \cref{eq:invariance_orthogonal}.

Using these properties, for all $\bx_t\in \mathcal{X}$ and $(R,\sigma)\in \mathrm{O}(m)\times \mathrm{S}_n$, we obtain
\begin{align}
    \tilde{q}((R, \sigma)\circ \bx_t) &= \int_{\mathcal{X}^{T-t}}\tilde{q}((R,\sigma)\circ \bx_{t:T})d{((R, \sigma)\circ \bx)}_{t+1:T}\label{eq:invariance_eq_1}\\
    &= \int_{((R, \sigma)^{-1}\circ \mathcal{X})^{T-t}}\tilde{q}((R,\sigma)\circ \bx_{t:T})\left\vert \text{det}\frac{\partial (R, \sigma)\circ \bx}{\partial \bx}\right\vert_{t+1:T} d\bx_{t+1:T}\label{eq:invariance_eq_2}\\
    &= \int_{\mathcal{X}^{T-t}}\tilde{q}((R,\sigma)\circ \bx_{t:T})d\bx_{t+1:T}\label{eq:invariance_eq_3}\\
    &=\int_{\mathcal{X}^{T-t}} \tilde{\pi}((R, \sigma )\circ \bx_{t:T})\frac{q((R, \sigma)\circ \bx_T)}{\tilde{\pi}((R, \sigma)\circ \bx_T)}d\bx_{t+1:T}\label{eq:invariance_eq_4}\\
    &=\int_{\mathcal{X}^{T-t}} \tilde{\pi}((R, \sigma )\circ \bx_t)\prod_{i=0}^{T-t-1}\tilde{\pi}((R, \sigma )\circ \bx_{t+i+1}|(R, \sigma )\circ \bx_{t+i})\frac{q((R, \sigma )\circ \bx_T)}{\tilde{\pi}((R, \sigma )\circ \bx_T)}d\bx_{t+1:T} \label{eq:invariance_eq_5} \\
    &= \int_{\mathcal{X}^{T-t}} \tilde{\pi}(\bx_t)\prod_{i=0}^{T-t-1}\tilde{\pi}(\bx_{t+i+1}| \bx_{t+i})\frac{q(\bx_T)}{\tilde{\pi}(\bx_T)}d\bx_{t+1:T} \label{eq:invariance_eq_6}\\
    &= \int_{\mathcal{X}^{T-t}} \tilde{q}(\bx_{t:T})d\bx_{t+1:T} \label{eq:invariance_eq_7}\\
    &=\tilde{q}(\bx_t).
\end{align}
\cref{eq:invariance_eq_2} follows from a change of variables in
\cref{eq:invariance_eq_1}. In \cref{eq:invariance_eq_3} we use the fact that 
$(R, \sigma)^{-1}\circ \mathcal{X} = \mathcal{X}$ and that 
$\bx \mapsto (R, \sigma)\circ \bx$ is an orthogonal transformation, so that
$\bigl\lvert \det \tfrac{\partial (R,\sigma)\circ \bx}{\partial \bx} \bigr\rvert = 1$. \cref{eq:invariance_eq_4} and \cref{eq:invariance_eq_7} follow from the definition
of the joint target distribution
$\tilde{q}(\bx_{t:T}) = \tilde{\pi}(\bx_{t:T})\, q(\bx_T)/\tilde{\pi}(\bx_T)$. Finally, \cref{eq:invariance_eq_6} uses the already
established invariance of $q(\bx_T)$, $\tilde{\pi}(\bx_t)$ and
$\tilde{\pi}(\bx_{t+1}| \bx_t)$.

Therefore, $\tilde{q}(\bx_t)$ is $\mathrm{O}(m)\times \mathrm{S}_n$-invariant, which yields the desired invariance of $V_*^t$ via \cref{eq:opt_val}.

\end{proof}

\begin{proposition}[$\mathrm{O}(m)$-Invariance of $V_\pi^t$ and $V_*^t$]
\label{proposition:O_m_invariance}
    Assume that the energy function is $\mathrm{O}(m)$–invariant as follows:
    \begin{align}
        E( R \cdot \bx) =  E(\bx) \quad \forall \bx\in \mathcal{X}, R \in \mathrm{O}(m) \nonumber
    \end{align}
    then both the value function of VGS and the optimal value function preserves $\mathrm{O}(m)$-invariance:
    \begin{align}
        V^{t}_{\pi}(R \cdot \bx_t) =  V^t_{\pi}(\bx_t), \quad 
        V^{t}_{*}(R \cdot \bx_t) =  V^t_{*}(\bx_t), \nonumber
    \end{align}
    for all $\bx_t\in \mathcal{X}$, $R \in \mathrm{O}(m)$.
\end{proposition}
\begin{proof}
It is straightforward to check that $\cdot:(R,\bx)\mapsto R\cdot \bx$ is an orthogonal action and that the isotropic zero-mean Gaussian is invariant under it. Therefore, the proof in Proposition \ref{proposition:invariance} applies after replacing the product action $\circ: ((R, \sigma), \bx) \mapsto ((R, \sigma) \circ \bx)$ with $\cdot:(R,\bx)\mapsto R\cdot \bx$. We omit the technical details.
\end{proof}

% \paragraph{Lower variance estimate of TD}

% \begin{align*}
%     r_t(\bx_t, \bx_{t+1}) = \tau\log\frac{\pi_{\phi}(\bx_{t+1}|\bx_t)}{\tilde{q}(\bx_t|\bx_{t+1}) } = \tau(-\frac{||\mu_t(\bx_t)||^2}{2\alpha_t^2s_t^2} + \frac{\mu_t(\bx_t)^T(\bx_{t+1}-\alpha_t\bx_t)}{\alpha_t^2s_t^2}-D\log{\alpha_t})
% \end{align*}
%%%%%%%%%%%%%%%%%%%%%%%%%%%%%%%%%%%%%%%%%%%%%%%%%%%%%%%%%%%%%%%%%%%%%%%%%%%%%%%
%%%%%%%%%%%%%%%%%%%%%%%%%%%%%%%%%%%%%%%%%%%%%%%%%%%%%%%%%%%%%%%%%%%%%%%%%%%%%%%

\section{Algorithms}
\begin{algorithm}[t]
   \caption{Learning Value Functions in VGS (TD(0))}
   \label{alg:vgs_training}
\begin{algorithmic}
   \STATE {\bf Input:} Energy $E(\bx)$, value  $V_{\phi}^t(\bx_t)$, schedules $\{\alpha_t\}_{t=0}^{T-1}$, $\{\sigma_t\}_{t=0}^{T-1}$.

   \FOR{$i=1$ {\bfseries to} $n_{iter}$}
   \STATE Sample trajectory $\bx_{0:T}\sim \pi_{\phi^-}(\bx_{0:T})$ \hfill // Alg. \ref{alg:vgs}
   \STATE Store tuples $\mathcal{D} = \{(\bx_{t}, \mu_{\phi^{-}}^{t}(\bx_t)) \}_{t=0}^{T}$
    \FOR{$(\bx_t, \mu_{\phi^-}^t(\bx_t))$ {\bfseries in} $\mathcal{D}$}
   % \STATE Resample next state $\bx'_{t+1} = \alpha_t\bx_{t} + \mu_{\phi^-}^t(\bx_t) +\sigma_t\epsilon_t$ \hfill //  Eq. (\ref{eq:pls})
   \STATE Compute the value target $\hat{V}^t(\bx_t)$ as $\hat V^t_{\text{TD}}(\bx_{t:t+1})$ for $t<T$, and as $\tilde{E}(\bx_T)$ for $t=T$
   \ENDFOR
   \FOR{$j=1$ {\bfseries to} $n_{update}$}
    \FOR{$(\bx_t, \_)$ {\bfseries in} $\mathcal{D}$}
   \STATE Optimize $\min_{\phi} ((V_{\phi}^{t}(\bx_{t}) - \hat{V}^t(\bx_{t}))^2$
   \ENDFOR
   \ENDFOR
   \STATE Update target value parameters $\phi^{-} \leftarrow \kappa \phi^- + (1-\kappa)\phi$
   \ENDFOR
\end{algorithmic}
\end{algorithm}
% \end{minipage}
% \end{wrapfigure}

% \begin{algorithm}[H]
%    \caption{Training EBM with VGS}
%    \label{alg:ebm-vgs}
% \begin{algorithmic}
%    \STATE {\bfseries Input:} EBM $E_\theta(\bx)$, Values $V_\phi^t(\bx_t)$, Dataset $\mathcal{D}=\{\bx_i\}_{i=1}^{N}$, Hyperparameter $\gamma>0$, Regularizer functional for the energy $Reg_\theta$.
%    % \REPEAT
%    % Generate sample using Algorithm 1
%    \FOR{each minibatch $\bx_i\sim\mathcal{D}$}
%    \STATE Sample $\bx^-\sim\pi_\phi(\bx)$ by \cref{alg:vgs}
%    % update energy
%    \STATE $\min_\theta E_\theta(\bx_i) - E_\theta(\bx^-) + \gamma Reg_\theta(\bx_i, \bx^-)$
%    % update value using Algorithm 2
%    \STATE Update $V_\phi^t$ using \cref{alg:vgs_training}.
%    \ENDFOR
%    % \UNTIL{$noChange$ is $true$}
% \end{algorithmic}
% \end{algorithm}

\begin{algorithm}[t]
   \caption{Computing off-policy TD($\lambda$) targets with an exploration policy}
   \label{alg:off_td_lambda}
\begin{algorithmic}
   \STATE {\bf Input:} Target value network $V_{\phi^-}^t(\bx_t)$, schedules $\{\alpha_t\}_{t=0}^{T-1}$, $\{\sigma_t\}_{t=0}^{T-1}$.
   \STATE Sampled trajectory $\bx_{0:T} \sim \pi_{\text{expl}}(\bx_{0:T})$ using $\mu^t_{\text{expl}}(\bx_t) = \mu_{\phi^-}^{t}(\bx_t),\ (\sigma_t)_{\text{expl}} = \eta \sigma_t$ \hfill // Alg. \ref{alg:vgs}
   \STATE Stored tuples $\mathcal{D} = \{(\bx_t, \mu_{\phi^-}^{t}(\bx_t))\}_{t=0}^{T}$
   \STATE 
   \STATE Initialize advantage estimate: $\hat{A}_T = 0$
   \FOR{$t = T\!-\!1, \dots, 0$}
       \STATE Resample $\bx'_{t+1}\sim \pi_{\phi^-}(\cdot|\bx_t)$ as $\bx'_{t+1} = \alpha_t \bx_t + \mu_{\phi^-}^{t}(\bx_t) + \sigma_t \epsilon_t$ \hfill // Eq.~(\ref{eq:pls})
       \STATE Compute TD error ${\delta}_t(\bx_t, \bx'_{t+1})$ \hfill // Eq.(\ref{eq:delta_t})
       \STATE Update advantage estimate
       $\hat{A}_t = \lambda  \frac{\pi_{\phi^-}(\bx_{t+1}|\bx_t)}{\pi_{\text{expl}}(\bx_{t+1}|\bx_t)} \hat{A}_{t+1} + \delta_t(\bx_t,\bx'_{t+1})$
       \STATE Compute TD target
       $\hat{V}^t_{\text{off-TD($\lambda$)}}(\bx_{t:T}) = V_{\phi^-}^t(\bx_t) + \hat{A}_t$ \hfill 
       // Eq.(\ref{eq:td_target_net})
   \ENDFOR
\end{algorithmic}
\end{algorithm}

The detailed algorithm for training the value function in VGS is presented in \cref{alg:vgs_training}. For clarity, we first present a minimal TD(0) version. In the more general setting where off-policy TD($\lambda$) targets are computed using an exploration policy, the TD target computation step in \cref{alg:vgs_training} is substituted with the procedure in \cref{alg:off_td_lambda}. To satisfy the boundary condition at $t=T$, $V^T_{\phi^-}(\bx_T)$ is replaced by $\tilde{E}(\bx_T)$ in the value target calculation. 

% \begin{wrapfigure}{r}{0.5\textwidth}
%   \begin{minipage}{0.48\textwidth}

\section{Experimental Setup}\label{appendix:experiment_appendix}
\subsection{Target Distributions}
\label{appendix:target_dist}
\textbf{GMM \citep{sendera2024improved}.}\quad We use a 2-dimensional Gaussian Mixture Model, with its density given by $\gamma(\bx) = \frac{1}{m} \sum_{i=1}^{m}\mathcal{N}(\bx;{\mu}_i,\Sigma_{i}).$, where $m = 25,({\mu}_i)_{i=1}^{25} = \{-10,-5,0,5,10\}\times\{-10,-5,0,5,10 \}   \subset \mathbb{R}^2$ and $(\Sigma_{i})_{i=1}^{25} = 0.3\mathbf{I} \subset \mathbb{R}^{2\times2}$. With these parameters, a 25-component Gaussian Mixture density with evenly separated modes is obtained. A well-trained sampler must be able to sample from all twenty five modes.

\textbf{Funnel \citep{neal2003funnel}.}\quad We use a 10-dimensional funnel distribution with its density given by $\gamma(\bx)=\mathcal{N}(x_1;{0},\sigma^2)\prod_{i=1}^{d}\mathcal{N}(x_i;{0},e^{x_1})$, where $\bx\in \mathbb{R}^d$. The specific parameters are $d = 10$ and $\sigma = 3$. 
The funnel distribution is known to be a  challenging distribution for testing MCMC methods, as the variance grows exponentially as the first dimension $x_1$ increases.

\textbf{ManyWell \citep{noe2019boltzmann}.}\quad We use a 32-dimensional manywell distribution with its density given by $\gamma(\bx) = \prod_{i=1}^{16}\mu(x_{2i-1}, x_{2i})$, where $\mu(x_{i}, x_{j}) = \exp (-x_{i}^{4} + 6x_{i}^{2}+0.5x_{i}-0.5x_{j}^{2})$ and $x \in \mathbb{R}^{32}$.

\textbf{DW-4 \citep{kohler2020equivariantflowsexactlikelihood}.}\quad  
This system consists of 4 particles in two dimensions, where each pair of particles interacts via a \textit{double-well} potential: $
\gamma^{\text{DW}}(\bx) = \frac{1}{2\tau} \sum_{i,j} \left[ a \left( d_{ij} - d_0 \right) + b \left( d_{ij} - d_0 \right)^2 + c \left( d_{ij} - d_0 \right)^4 \right],$
where \(d_{ij}\) denotes the Euclidean distance between particles \(i\) and \(j\).  
These pairwise interactions induce multiple metastable states in the system.  
 We use the same parameters as in prior works~\citep{midgley2023se, akhound-sadegh2024iterated, he2024training}, setting \(d_0 = 4\), \(a = 0\), \(b = -4\), \(c = 0.9\), and \(\tau = 1\). We split the MCMC samples from \cite{klein2023equivariantflowmatching} into validation and test sets, which serve as our reference ground truth samples. We use 1,000 samples for validation dataset, and 10,000 for test dataset.

\textbf{LJ-13/LJ-55 \citep{kohler2020equivariantflowsexactlikelihood}.}\quad  
We consider systems of 13 and 55 particles in three dimensions, where particles interact via the \textit{Lennard-Jones} (LJ) potential: $\gamma^{\text{LJ}}(\bx) = \frac{\epsilon}{2\tau} \sum_{i,j} \left[ \left( \frac{r_m}{d_{ij}} \right)^{12} - 2 \left( \frac{r_m}{d_{ij}} \right)^6 \right],$
with \(d_{ij}\) denoting the Euclidean distance between particles \(i\) and \(j\).  
In this setting, the energy and its gradient become unbounded as \(d_{ij} \to 0\), making the optimization landscape challenging.  
To prevent particle dissociation, we follow prior works~\citep{midgley2023se, akhound-sadegh2024iterated, he2024training} and add a harmonic potential centered at the system's center of mass (CoM): $
\gamma^{\text{osc}}(\bx) = \frac{1}{2} \sum_i \| x_i - x_{\text{CoM}} \|^2.$
The same works also set \(\epsilon = 1\), \(r_m = 1\), and \(\tau = 1\), which we adopt in our experiments.We split the MCMC samples from \cite{klein2023equivariantflowmatching} into validation and test sets, which serve as our reference ground truth samples. We use 1,000 samples for validation dataset, and 10,000 for test dataset.

\subsection{Performance Metrics}
\label{appendix:metrics}

\textbf{Total Variation Distance (TVD-D, E).}\quad The total variation distance (TVD) between two distributions $P$ and $Q$ is defined as $
\text{TVD}(P, Q) = \frac{1}{2} \int_{\mathcal{X}} |P(x) - Q(x)| \, dx.$
However, accurate computation of TVD directly in high-dimensional data space is intractable in practice.  
Instead, we project the data to physically meaningful one-dimensional statistics—interatomic distances and total energies—and compute TVD on the resulting distributions.  
We refer to these metrics as TVD-D (for distances) and TVD-E (for energies), respectively.  
Each is approximated using histogram-based probability distributions, with the number of bins determining the discretization scale.  
We use 200 bins for all experiments.

\textbf{Log Partition Function Error ($\Delta \log Z_r, \Delta \log Z_f$).}\quad 
The log partition function error is defined as the difference between the target density’s true log partition function and its estimated value, denoted as, $\Delta \log Z = |\log Z - \log \hat{Z}|$ \citep{blessing2024elboslargescaleevaluationvariational}.
The estimate $\hat{Z}$ is computed by using samples from either the trained distribution ($\pi_\phi(\mathbf{x})$) or target ($q(\mathbf{x})=\gamma(\mathbf{x})/Z$) distribution.
Depending on the distribution used, the estimates are named as reversed estimate ($\hat{Z}_r$) or forward estimate ($\hat{Z}_f$) respectively.
Each estimates and its corresponding log partition function error is defined as
\begin{align*}
    \hat{Z}_r = \mathbb{E}_{\substack{\mathbf{x}_0 \sim \mathcal{N}(0, \sigma_{\text{init}}^2 I) \\ \mathbf{x}_{t+1}\sim \pi_{\phi}(\mathbf{x}_{t+1}|\mathbf{x}_{t})}} \left[ \frac{\gamma(\mathbf{x}_T)\prod_{t=0}^{T-1}\tilde{q}(\mathbf{x}_t|\mathbf{x}_{t+1})}{p(\mathbf{x}_0)\prod_{t=0}^{T-1}\pi_{\phi}(\mathbf{x}_{t+1}|\mathbf{x}_t)} \right] \quad& \Delta \log Z_r = \left| \log Z - \log \hat{Z}_r \right| \\
    \hat{Z}_f = 1/\mathbb{E}_{\substack{\mathbf{x}_T\sim q(\mathbf{x}_T) \\ \mathbf{x}_{t-1}\sim\tilde{q}(\mathbf{x}_{t-1}|\mathbf{x}_t)}}\left[ \frac{p(\mathbf{x}_0)\prod_{t=0}^{T-1}\pi_{\phi}(\mathbf{x}_{t+1}|\mathbf{x}_t)}{\gamma(\mathbf{x}_T)\prod^{T-1}_{t=0}\tilde{q}(\mathbf{x}_t|\mathbf{x}_{t+1})} \right] \quad& \Delta \log Z_f = \left| \log Z - \log \hat{Z}_f \right|.
\end{align*}

\textbf{Wasserstein-2 Distance ($\mathcal{W}^2$).}\quad The 2-Wasserstein distance between two distributions is defined as \( \mathcal{W}_2(P, Q) = \left( \inf_{\pi} \int \pi(x, y) d(x, y)^2 \,dx \,dy \right)^{\frac{1}{2}} \),
where \(\pi \) is the transport plan with marginals constrained to \(P\) and \(Q\) respectively \citep{akhound-sadegh2024iterated}. We use the Euclidean distance \(||x-y||_2\) for \(d(x,y)\) and calculate the 2-Wasserstein distance between generated samples from the sampler and the ground truth data. For the actual computation, we use the Python Optimal Transport Library \citep{flamary2021pot}.

\begin{table}[t]
  \setlength\tabcolsep{3pt}
  \centering
  % \small
  \caption{Hyperparameters for VGS used in $n$-body system experiments}
  \label{tab:vgs-hparams_n_body}
  % \scriptsize
  \resizebox{1\linewidth}{!}{
  \begin{tabular}{@{} ccc*{17}{c}@{}}
    \toprule
    \textbf{Target} 
    & $T$ & $n_{iter}$ & $n_{update}$ & $\lambda$
    & $\sigma_0^2$ & $\sigma^2_{T-1}$ & var schedule & $\alpha_t$ & $\sigma_{\text{init}}$ & $\eta$ & $\kappa$ & $\text{lr}$
    & Clip $\tilde{E}(\bx)$ & Clip $\hat{A}_t$ & Hidden dim 
     \\
    \midrule
    \addlinespace
    {\textbf{DW-4}} 
    &  50 & 50000 & 3 & 0 & 0.2 & 0.001 & quad & 1 (VE) & 0 
    & 1.2 & 0.9 & $1e\text{-}5$ 
    & - & - & 256 \\
    \addlinespace
    \hline
    \addlinespace
    {\textbf{LJ-13}} 
    & 100 & 50000 & 3 & 0.9 & 0.05 & 0.0001 & exp & 1 (VE) & 0
    & 1.2 & 0.98 & $1e\text{-}5$ 
    & $1e4$ & 100 & 512\\
    \addlinespace
    \hline
    \addlinespace
    {\textbf{LJ-55}} 
    & 100 & 50000 & 3 & 0.9 & 0.03 & 0.0001 & quad & 1 (VE) & 0
    & 1.2 & 0.95 & $1e\text{-}5$ 
    & $1e6$ & 100 & 1024\\
    \addlinespace
    \bottomrule
  \end{tabular}}
\end{table}

\subsection{Experimental Details}
\label{appendix:exp_setup}

\subsubsection{Sampling from \texorpdfstring{$n$}{n}-Body System Experiments}\label{appendix:particle_detail}

For the $n$-body system experiments, we conducted 3 independent runs with different random seeds for each setting. All reported metrics are the mean and standard deviation over these 3 runs. Each experiment was conducted on a single RTX 3090 GPU or RTX 4090 GPU with 24 GB of memory. Below, we provide detailed implementation and hyperparameters for each algorithm used.

We applied early stopping during training based on 1,000 validation samples and saved the checkpoint with the lowest TVD-D score, following the protocol in \cite{he2024training}. For evaluation, we report metrics computed on 10,000 samples generated from the final checkpoint and compare them against the test set. Since direct sampling from the target potentials is infeasible, we use the validation and test sets provided in the official iDEM repository at \url{https://github.com/jarridrb/DEM} \citep{akhound-sadegh2024iterated}, which is adopted from \cite{klein2023equivariantflowmatching}. As the original validation set contains 10,000 samples, we randomly select and fix a subset of 1,000 samples for early stopping. We provide implementation details and hyperparameter settings for each algorithm below.

\textbf{Value Gradient Sampler.}\quad The hyperparameters used for VGS are summarized in \cref{tab:vgs-hparams_n_body}.  As described in \cref{alg:vgs_training}, $n_{iter}$ denotes the total number of training iterations, and $n_{update}$ controls how many times the rollout samples $\mathcal{D}$ is reused for gradient updates. We use a batch size of 512 during the trajectory sampling phase and 2048 for TD updates. We use the Adam optimizer \citep{kingma2017adammethodstochasticoptimization} for gradient update.

We use the IMLP value-function architecture in our main experiments as it is more computationally efficient than IGNN. IMLP takes the sorted pairwise distances $\{d_{ij}\}_{i<j\le n}$ as input; for the LJ experiments, we instead use their inverses, motivated by the analytic form of the LJ potential. To prevent numerical divergence at $d = 0$, a small constant is added. The network is a four-layer MLP with ReLU activations. We encode the time step with a 128-dimensional sinusoidal embedding and project it to the hidden dimension via a fully connected layer. We adjust model complexity via the hidden dimension, which we report across experiments in the table.

The variance schedule $\{\sigma_t^2\}_{t=0}^{T-1}$ is defined by choosing the endpoints $\sigma_0^2$ and $\sigma_{T-1}^2$, then interpolating intermediate values according to a predefined rule (see ``var schedule'' in the table). In all experiments we use a variance-exploding(VE) reference process $\tilde{\pi}$ initialized from a Dirac delta, i.e., $\alpha_t=1$ and $\sigma_{\text{init}}=0$.

The hyperparameter $\eta$ determines the noise scale of the exploration policy $(\sigma_t)_{\text{expl}}=\eta\sigma_t$, and thus the exploration level during sampling. For simplicity, we do not decay $\eta$ during training. The parameter $\kappa$ is the momentum coefficient for the exponential moving average used to update the target value network.

We find that clipping the terminal rewards $\tilde{E}(\bx)$ and the advantage estimates $\hat{A}_t$ (defined in \cref{alg:off_td_lambda}) improves training stability. The clipping thresholds are reported in the table as ``Clip $\tilde{E}(\bx)$'' and ``Clip $\hat{A}_t$''.

The LJ potential is particularly sensitive to sample noise, so during evaluation we omit the final diffusion step by setting $\sigma_{T-1}=0$.

    % \multirow{2}{*}{\textbf{DW-4}} 
    % & \textbf{VGS} & 30 & 3000 & 3 & 0 & 0.1 & 0.0001 & quad & 1.2 & 0 & $1e\text{-}5$ & $1e\text{-}3$ & - & - & 256 \\
    % & \textbf{+TD($\lambda$)} & 30 & 3000 & 3 & 0.9 & 0.1 & 0.0001 & quad & 1.2 & 0 & $1e\text{-}5$ & $1e\text{-}3$ & - & - & 256 \\

\textbf{Baseline Methods.}\quad We conduct baseline experiments using the following repositories: FAB with an $\mathrm{SE}(3)$-augmented coupling flow architecture \citep{midgley2023se, midgley2023flow} from \url{https://github.com/lollcat/se3-augmented-coupling-flows}, iDEM \citep{akhound-sadegh2024iterated} from \url{https://github.com/jarridrb/DEM}, DiKL \citep{he2024training} from \url{https://github.com/jiajunhe98/DiKL}, PIS and DDS \citep{berner2022optimal, richter2024improved} from \url{https://github.com/juliusberner/sde_sampler}, and GFlowNets (GFN-DB, GFN-TB, GFN-SubTB) from \url{https://github.com/GFNOrg/gfn-diffusion}.

For all baselines, we apply early stopping based on validation TVD-D, following the same protocol used for VGS. A fixed validation set of 1,000 samples and a test set of 10,000 samples are used across all experiments (\cref{appendix:target_dist}). We use the default hyperparameters provided in each repository unless otherwise specified, with three exceptions.

First, FAB and DiKL do not provide default configurations for LJ-55, so we adopt their LJ-13 hyperparameters. Because their equivariant networks inherently scale computation with the number of particles, we do not explicitly increase the hidden dimensions or number of layers.

Second, to address out-of-memory issues in certain experiments, we selected the largest batch size our hardware allows. For instance, the batch size for FAB is reduced from 1024 to 128 on LJ-13, and to 8 on LJ-55. For DiKL on LJ-55, the batch size is decreased from 256 to 64.

Third, since PIS, DDS, and GFlowNets did not originally include particle system experiments, we replaced their MLP-based policy networks with EGNNs, following the default configurations in iDEM \citep{akhound-sadegh2024iterated}. We used an EGNN with a hidden dimension of 128, utilizing 3 hidden layers for DW-4 and 5 hidden layers for LJ-13 and LJ-55.

\subsubsection{Sampling from Synthetic Distribution Experiments}\label{appendix:synthetic_details}
For the synthetic distribution experiments, we conducted 3 independent runs with different seeds for each experiment. All reported metrics are the averages of the 3 runs, along with their standard deviation. Every experiment is performed on a single NVIDIA RTX 3090 (24GB) GPU. To compute the metrics, we use 2,000 randomly sampled data from a target density and another 2,000 randomly sampled data from a learned sampler to evaluate the metrics. Below, we provide detailed implementation and hyperparameters for each algorithm used.

\textbf{Value Gradient Sampler.}\quad The hyperparameters used for VGS are summarized in \cref{tab:vgs-hparams}. Consistent with the $n$-body system experiments, we use a batch size of 512 during trajectory sampling and 2048 for training. In our experiments with buffer, we adopt the buffer design from \cite{sendera2024improved}, which maintains a prioritized mixture of MCMC and policy-generated samples based on sample energy. Following their approach, we alternate training VGS on forward and backward trajectories.

\begin{table}[t]
  \setlength\tabcolsep{3pt}
  \centering
  % \small
  \caption{Hyperparameters for VGS used in synthetic distribution experiments}
  \label{tab:vgs-hparams}
  % \scriptsize
  \resizebox{1\linewidth}{!}{
  \begin{tabular}{@{} cc*{17}{c}@{}}
    \toprule
    \textbf{Target} & \textbf{Methods} 
    & $T$ & $n_{\text{iter}}$ & $n_{\text{update}}$ & $\lambda$
    & $\sigma^2_0$ & $\sigma^2_{T-1}$ & var schedule
    &$\alpha_t$ & $\sigma_{\text{init}}$& $\eta$ & $\kappa$ & $\text{lr}$ & Clip $E(\bx)$ & Clip $\hat{A}_t$ & Hidden dim \\
    \midrule
    \addlinespace
    & \textbf{VGS} & 50 & 5000 & 3 & 0 & 0.1 & 0.1 & const & 1 (VE) & 0 & 1.2 & 0.98 & $1e\text{-}4$ & - & - & 256 \\
    \addlinespace
    \textbf{GMM} & \textbf{+TD($\lambda$)} & 50 & 5000 & 3 & 0.9 & 0.1 & 0.1 & const & 1 (VE) & 0 & 1.2 & 0.98 & $1e\text{-}4$ & - & - & 256 \\
    \addlinespace
    & \textbf{+Buffer} & 50 & 5000 & 1 & 0 & 0.1 & 0.1 & const & 1 (VE) & 0 & 1.1 & 0.98 & $1e\text{-}4$ & - & - & 256 \\
    \addlinespace
    \hline
    \addlinespace
    & \textbf{VGS} & 100 & 20000 & 3 & 0 & 0.03 & 0.0001 & quad & 1 (VE) & 0 & 1.0 & 0.98 & $1e\text{-}5$ &  $1e3$ & 100 & 256\\
    \addlinespace
    \textbf{Funnel} & \textbf{+TD($\lambda$)} & 100 & 20000 & 3 & 0.9 & 0.03 & 0.0001 & quad & 1 (VE) & 0 & 1.0 & 0.98 & $1e\text{-}5$ &  $1e3$ & 100 & 256 \\
    \addlinespace
    & \textbf{+Buffer} & 100 & 20000 & 1 & 0 & 0.03 & 0.0001 & quad & 1 (VE) & 0 & 1.0 & 0.98 & $1e\text{-}5$ &  $1e3$ & 100 & 256 \\
    \addlinespace
    \hline
    \addlinespace
    & \textbf{VGS} & 100 & 20000 & 3 & 0 & 0.03 & 0.0001 & quad & 1 (VE) & 0 & 1.2 & 0.98 & $1e\text{-}5$ & $1e3$ & 100 & 512\\
    \addlinespace
    \textbf{ManyWell} & \textbf{+TD($\lambda$)} & 100 & 20000 & 3 & 0.9 & 0.03 & 0.0001 & quad & 1 (VE) & 0 & 1.2 & 0.98 & $1e\text{-}5$ & $1e3$ & 100 & 512 \\
    \addlinespace
    & \textbf{+Buffer} & 100 & 20000 & 1 & 0 & 0.03 & 0.0001 & quad & 1 (VE) & 0 & 1.1 & 0.98 & $1e\text{-}5$ & $1e3$ & 100 & 512  \\
    \addlinespace
    \bottomrule
  \end{tabular}}
\end{table}

\textbf{SMC-based Methods.}\quad The implementation of SMC-based method is based on \url{https://github.com/google-deepmind/annealed_flow_transport}. The implementation of SMC variants, AFT \citep{arbel2021annealedflowtransportmonte} and CRAFT \citep{matthews2023continualrepeatedannealedflow} are provided together. For GMM, an initial gaussian distribution with $\sigma = 10.0$ is used to ensure suitable exploration in the range $[-10,10] \times [-10,10] \in \mathbb{R}^2$ where the modes exists. The default hyper-parameter for the funnel distribution and the manywell distribution is provided in the code. For all the experiments, the resample threshold of 0.3 is used. Additional hyperparameters are reported in \cref{tab:smc_hparams}.

% For SMC, we used 2000 particles and 128
% annealing steps. We used resampling with a threshold of 0.3. We used one Hamiltonian Monte Carlo (HMC) step per temperature with 10 leapfrog steps.
% Continual Repeated Annealed Flow Transport (CRAFT/AFT). As AFT and CRAFT build on Sequential Monte Carlo
% (SMC), we employed the same SMC specifications detailed above. Notably, we found that employing simpler flows in
% conjunction with a greater number of temperatures yielded superior and more robust performance compared to the use of
% more sophisticated flows such as RealNVP or Neural Spline Flows. Consequently, we opted for 128 temperatures, utilizing
% diagonal affine flows as the transport map. Specifically for AFT, we determined that 400 iterations per temperature were
% sufficient to achieve converged training results. For CRAFT and SNF, we found that a total of 3000 iterations provided
% satisfactory convergence during training. For all methods, we use 2000 particles for training and testing and tune the learning
% rate and the scale of the initial proposal distribution π0 as shown in Table 8.

\begin{table}[t]
\centering
\caption{Hyperparameters for SMC-based Methods used in synthetic distribution experiments}
\label{tab:smc_hparams}
\begin{small}
\begin{tabular}{@{} l c c c @{}}
\toprule
\textbf{Methods / Parameters} & \textbf{GMM} & \textbf{Funnel} & \textbf{ManyWell} \\
\midrule
\textbf{SMC} & & & \\
Initial $\sigma$ & 10.0 & 1.0 & 12.0\\
HMC stepsize ($\beta\le0.5$) & 0.2 & 0.01 & 0.01 \\
HMC stepsize ($\beta>0.5$) & 0.2 & 0.1 & 0.1 \\
\midrule
\textbf{AFT} & & & \\
Initial $\sigma$ & 9.0 & 1.0 & 10.0 \\
Learning Rate & $1e\text{-}3$ & $1e\text{-}3$ & $1e\text{-}3$ \\
\midrule
\textbf{CRAFT} & & & \\
Initial $\sigma$ & 9.0 & 1.0 & 10.0 \\
Learning Rate & $1e\text{-}2 $& $1e\text{-}3$ & $1e\text{-}3$ \\
\bottomrule
\end{tabular}
\end{small}
\end{table}

\textbf{SDE-based Methods.}\quad For SDE-based methods, we follow the implementation of the official repository of the DIS \cite{berner2022optimal, richter2024improved}. The implementation of 
PIS \citep{zhang2022path}, DIS \citep{berner2022optimal} and DDS \citep{vargas2023denoising, richter2024improved} are all available. 
 The PIS, DIS and DDS is trained using the default setting provided by the authors using the log-variance training objective \citep{richter2024improved}. Specifically, we used timesteps of $T=200$ for PIS, DIS and $T=256$ for DDS. We trained all models for 30,000 iterations with the learning rate value of $5\mathrm{e}\text{-}3$. The Fourier MLP architecture \citep{zhang2022path} was used, which is fundamentally a MLP architecture with sinusoidal time step embeddings. The network is composed of 4 layers with the hidden dimension of size 64 and the time step embedding dimension of size 64.

% \textbf{DSM-based Methods}: For DSM-based methods, we adopt the official repository as provided by their respective authors, DiKL \cite{he2024training} and iDEM \cite{akhound-sadegh2024iterated}.
% For target densities supported in each work's repository, we did not make any changes to configurations and hyperparameters.
% In the case of unsupported target densities, we used the configuration and hyperparameters from the 40 Mixture of Gaussian settings, changing only necessary configurations such as input dimensions.

\textbf{GFN Methods.}\quad
For the GFN baselines, we use the official implementation available at \url{https://github.com/GFNOrg/gfn-diffusion}
, corresponding to \cite{sendera2024improved}. This repository provides implementations of prior GFN studies on training diffusion samplers, which we refer to as GFN-TB \citep{lahlou2023theory}, GFN-SubTB \citep{zhang2023diffusion}, and GFN-TB-Imp \citep{sendera2024improved} in \cref{tab:main}. For all baselines, we adopt the default architectural configurations and training techniques provided by the authors, including linearly decaying exploration noise during training.

Specifically, for GFN-TB \citep{lahlou2023theory}, we employ the trajectory balance (TB) loss as described in the original work. For GFN-SubTB \citep{zhang2023diffusion}, we use the sub-trajectory balance (SubTB) loss with $\lambda = 2$, and incorporate an energy-informed flow together with a score-informed policy network, following the architectural design in \cite{zhang2023diffusion}. For GFN-TB-Imp \citep{sendera2024improved}, we apply the log-variance loss, a variant of the TB loss, and similarly use a score-informed policy network. In addition, we employ a prioritized replay buffer combined with local search, an improved off-policy strategy introduced in \cite{sendera2024improved}. Lastly, although GFN-DB \citep{bengio2023gflownetfoundations} was not originally evaluated with diffusion samplers, we include it for completeness. We use its basic architecture and a training procedure consistent with the other baselines, tuning only the learning rates.

\subsubsection{Ablation Experiments}

\textbf{VGS Architecture Ablation (\cref{fig:lj13_histogram}).}\quad
Experiments were conducted on the LJ-13 potential. The histogram was plotted using 10,000 samples. Details of the test samples are provided in \cref{appendix:particle_detail}.

For invariant MLP(IMLP), we used the same hyperparameters as in the main results (\cref{tab:particle_tab}). For non-invariant MLP, we kept the setup identical to IMLP but replaced the input from the sorted pairwise distances $\{d_{ij}\}_{i<j\le n}$ to the raw $39$-dimensional coordinate vector $\bx$. To isolate architectural effects, non-invariant MLP was also trained and evaluated in the zero-mean space $\mathcal{X}$.

For IGNN, we used three message-passing layers with hidden dimension $64$; the final node embeddings were sum-pooled and passed to a two-layer MLP with hidden dimension $64$. Due to computational constraints, we reduced the batch size to $64$ for both sampling and training. Due to altered training dynamics under this smaller batch and different architecture, we set the learning rate to $1\mathrm{e}{-4}$ and the momentum coefficient $\kappa$ to $0.9$.

\textbf{TD(0) vs. TD($\lambda$) Comparison (\cref{fig:td_lambda_analysis}).}\quad We provide the experimental details for  \cref{fig:td_lambda_analysis} which demonstates the benefit of TD($\lambda$) over TD(0). The experiment was conducted on the funnel distribution and the experimental details are given as follows. 

The left panel shows sample efficiency experiment demonstrating that TD($\lambda$) converges faster than TD(0). This behavior is mirrored by their GFN counterparts, GFN-SubTB and GFN-DB, respectively. For all methods, we searched over the learning rate grid $\{1\mathrm{e}\text{-}2, 1\mathrm{e}\text{-}3, 1\mathrm{e}\text{-}4, 1\mathrm{e}\text{-}5\}$. For each method, the largest value that did not cause divergence was selected. All hyperparameters were kept identical to those used in \cref{tab:vgs-hparams}.

The right panel of \cref{fig:td_lambda_analysis} shows how sampling steps effects the sampler performance, demonstrating that TD($\lambda$) achieves better performance than TD(0) for large timesteps $T$. This trend is likewise mirrored in their GFN counterparts. For VGS, we used a quadratic variance schedule and adjusted the endpoints $\sigma_0^2$ and $\sigma_{T-1}^2$, to ensure that the total injected noise remains constant regardless of $T$. For GFN, we adopted a constant variance schedule consistent with the main experimental settings (\cref{tab:main}), while similarly ensuring that the total injected noise was preserved. All other hyperparameters were kept identical to the main experiment.

\end{document}